\documentclass{article} 
\usepackage{iclr2023_conference,times}

\usepackage{amsmath,amsfonts,bm}

\def\eqref#1{equation~\ref{#1}}

\def\1{\bm{1}}

\DeclareMathAlphabet{\mathsfit}{\encodingdefault}{\sfdefault}{m}{sl}
\SetMathAlphabet{\mathsfit}{bold}{\encodingdefault}{\sfdefault}{bx}{n}

\usepackage{url}
\usepackage{graphicx}
\usepackage{amsmath,amssymb} 
\usepackage{cite}
\usepackage{color}
\usepackage{wrapfig}
\usepackage{epsfig}
\usepackage{graphicx}
\usepackage{amsmath}
\usepackage{amssymb}
\usepackage{multirow}
\usepackage{booktabs}
\usepackage{listings}
\usepackage{algorithm}
\usepackage{algorithmic}
\usepackage{arydshln}
\usepackage{threeparttable}
\usepackage{colortbl}
\usepackage{enumitem}
\usepackage{siunitx}
\usepackage{placeins}
\usepackage{bm}
\usepackage[numbers,sort&compress]{natbib}
\usepackage{array}
\usepackage{caption}
\usepackage{dblfloatfix}
\usepackage{amssymb}
\usepackage{pifont}

\usepackage[export]{adjustbox}

\usepackage{subcaption}

\definecolor{mygray}{gray}{.9}
\definecolor{ggray}{RGB}{127,127,127}
\definecolor{reda}{RGB}{192,0,0}
\definecolor{redb}{RGB}{217,148,143}
\definecolor{myyellow}{RGB}{190,144,0}
\definecolor{mygreen}{RGB}{80,100,40}
\definecolor{myblue}{RGB}{30,90,100}

\definecolor{mygreen2}{RGB}{80,100,40}  
\newcommand{\reshl}[2]{
{#1}\fontsize{7.5pt}{1em}\selectfont{\color{mygreen2}{$\uparrow$\text{#2}}}
}
\newcommand{\reshx}[2]{
{#1}\fontsize{7.5pt}{1em}\selectfont{\color{mygreen2}{$\uparrow$\textbf{#2}}}
}

\newcommand{\etal}{\textit{et al}.}
\newcommand{\ie}{\textit{i}.\textit{e}.}
\newcommand{\eg}{\textit{e}.\textit{g}.}

\makeatletter
\newcommand{\thickhline}{%
    \noalign {\ifnum 0=`}\fi \hrule height 1pt
    \futurelet \reserved@a \@xhline
}

\usepackage[pagebackref=true,breaklinks=true,letterpaper=true,colorlinks,bookmarks=false]{hyperref}

\title{$_{\!\!\!\!\!}$Visual$_{\!}$ Recognition$_{\!}$ with$_{\!}$ Deep$_{\!}$ Nearest$_{\!}$ Centroids$_{\!\!\!\!}$}

\author{Wenguan Wang$\rm^1$\thanks{Equal contribution}~~\footnotemark[2]~~, Cheng Han$\rm ^2$\footnotemark[1], Tianfei Zhou$\rm^3$\footnotemark[1] \& Dongfang Liu$\rm ^2$\thanks{Corresponding author} \\
CCAI, Zhejiang University$\rm^1$, Rochester Institute of Technology$\rm ^2$ \& ETH Zurich$\rm^3$ \\
}

\iclrfinalcopy 
\begin{document}

\maketitle

\vspace{-0.1cm}
\begin{abstract}
$_{\!}$We$_{\!}$ devise$_{\!}$ \underline{d}eep$_{\!}$ \underline{n}earest$_{\!}$ \underline{c}entroids$_{\!}$ (DNC),$_{\!}$ a$_{\!}$ conceptually$_{\!}$ elegant$_{\!}$ yet$_{\!}$ surprisingly$_{\!}$~ef- fective network for large-scale visual recognition, by revisiting Nearest~Centroids, one$_{\!}$ of$_{\!}$ the$_{\!}$ most$_{\!}$ classic$_{\!}$ and$_{\!}$ simple$_{\!}$ classifiers.$_{\!}$ Current$_{\!}$ deep$_{\!}$ models$_{\!}$ learn$_{\!}$ the$_{\!}$~classi- fier$_{\!}$ in$_{\!}$ a$_{\!}$ \textit{fully$_{\!}$ parametric}$_{\!}$ manner,$_{\!}$ ignoring$_{\!}$ the$_{\!}$ latent$_{\!}$ data$_{\!}$ structure$_{\!}$ and$_{\!}$~lacking 
explainability.$_{\!}$ DNC$_{\!}$ instead$_{\!}$ conducts$_{\!}$ nonparametric,$_{\!}$ case-based$_{\!}$ reasoning;$_{\!}$ it$_{\!}$ utilizes sub-centroids$_{\!}$ of$_{\!}$ training$_{\!}$ samples$_{\!}$ to$_{\!}$ describe$_{\!}$ class$_{\!}$ distributions$_{\!}$ and$_{\!}$ clearly$_{\!}$ ex- plains~the classification as the proximity of test data to the class sub-centroids in$_{\!}$ the$_{\!}$ feature$_{\!}$ space.$_{\!}$ Due$_{\!}$ to$_{\!}$ the$_{\!}$ distance-based$_{\!}$ nature,$_{\!}$ the$_{\!}$ network$_{\!}$ output$_{\!}$ dimensionality$_{\!}$ is$_{\!}$ flexible,$_{\!}$ and$_{\!}$ all$_{\!}$ the$_{\!}$ learnable$_{\!}$ parameters$_{\!}$ are$_{\!}$ \textit{only}$_{\!}$ for$_{\!}$ data$_{\!}$ embedding.$_{\!}$ That$_{\!}$ means$_{\!}$ all$_{\!}$ the$_{\!}$ knowledge$_{\!}$ learnt$_{\!}$ for$_{\!}$ ImageNet$_{\!}$ classification$_{\!}$ can$_{\!}$ be$_{\!}$ \textit{completely}$_{\!}$ transferred for pixel recognition learning, under the ``pre-training and fine-tuning'' paradigm. Apart from its nested simplicity and intuitive decision-making$_{\!}$ mechanism,$_{\!}$ DNC$_{\!}$ can$_{\!}$ even$_{\!}$ possess$_{\!}$ \textit{ad-hoc}$_{\!}$ explainability$_{\!}$ when the sub-centroids are selected as ac- tual$_{\!}$ training$_{\!}$ images$_{\!}$ that$_{\!}$ humans$_{\!}$ can$_{\!}$ view$_{\!}$ and$_{\!}$ inspect.$_{\!}$ Compared$_{\!}$ with$_{\!}$ parametric$_{\!}$ counterparts,$_{\!}$ DNC$_{\!}$ performs$_{\!}$ better$_{\!}$ on$_{\!}$ image$_{\!}$ classification$_{\!}$ (CIFAR-10,$_{\!}$ CIFAR-100,$_{\!}$ ImageNet)$_{\!}$ and greatly boosts pixel recognition  (ADE20K, Cityscapes) with improved transparency,$_{\!}$ using$_{\!}$ various$_{\!}$ backbone$_{\!}$ network$_{\!}$ architectures$_{\!}$ (ResNet,$_{\!}$ Swin)$_{\!}$ and$_{\!}$ segmentation$_{\!}$ models$_{\!}$ (FCN,$_{\!}$ DeepLab$_{\text{V3}}$, Swin). Our code is available at \href{https://github.com/ChengHan111/DNC}{DNC}. 

\end{abstract}

\vspace{-0.2cm}
\section{Introduction}
\begin{figure*}[t]
  \centering
        \vspace{-12pt}
      \includegraphics[width=1 \linewidth]{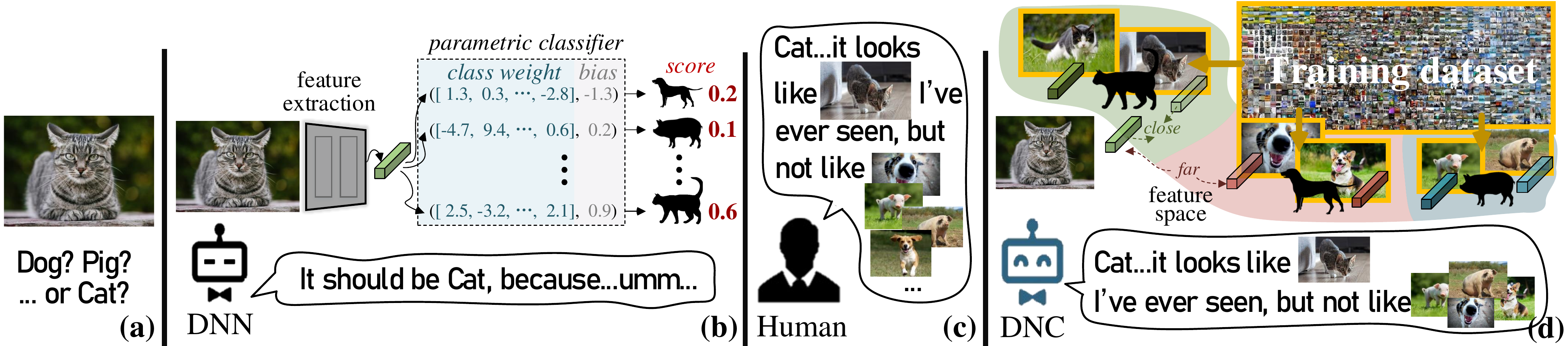}
      \vspace{-15pt}
\caption{(b) Prevalent visual recognition models \protect\includegraphics[scale=0.15,valign=c]{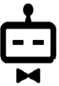}, built upon parametric softmax classifiers, have$_{\!}$ a$_{\!}$ few$_{\!}$ limitations,$_{\!}$ such$_{\!}$ as$_{\!}$ their$_{\!}$ non-transparent$_{\!}$ decision-making$_{\!}$ process.$_{\!}$~(c) Humans \protect\includegraphics[scale=0.12,valign=c]{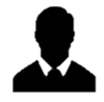} can use past cases as models when solving new problems$_{\!}$~\cite{newell1972human,aamodt1994case} (\eg, comparing$_{\!}$ \protect\includegraphics[scale=0.13,valign=c]{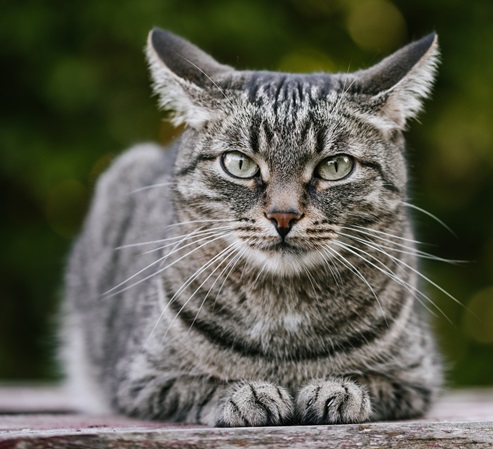}$_{\!}$ with$_{\!}$ a$_{\!}$ few$_{\!}$ familiar/exemplar$_{\!}$ animals$_{\!}$  for$_{\!}$ categorization).$_{\!}$ (d) DNC \protect\includegraphics[scale=0.15,valign=c]{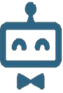} makes$_{\!}$ classification$_{\!}$ based$_{\!}$ on$_{\!}$ the$_{\!}$ similarity$_{\!}$ of$_{\!}$ \protect\includegraphics[scale=0.13,valign=c]{figs/cat}$_{\!}$ to$_{\!}$ class$_{\!}$ sub-centroids$_{\!}$ (representative$_{\!}$ training$_{\!}$ examples)$_{\!}$ in$_{\!}$
the$_{\!}$ feature$_{\!}$ space.$_{\!}$ The$_{\!}$ class sub-centroids$_{\!}$ are$_{\!}$ vital$_{\!}$ for$_{\!}$ capturing underlying data structure, enhancing interpretability, and boosting recognition. }
\label{fig:idea}
\vspace{-12pt}
\end{figure*}

\vspace{-0.25cm}
Deep$_{\!}$ learning$_{\!}$ models,$_{\!}$ from$_{\!}$ convolutional$_{\!}$ networks$_{\!}$ (\eg,$_{\!}$ VGG$_{\!}$~\cite{Simonyan15},$_{\!}$ ResNet$_{\!}$~\cite{he2016deep})$_{\!}$ to$_{\!}$ Transformer-based$_{\!}$\\
\noindent architectures$_{\!}$ (\eg,$_{\!}$ Swin$_{\!}$~\cite{liu2021swin}),$_{\!}$ push$_{\!}$ forward$_{\!}$ the$_{\!}$ state-of-the-art$_{\!}$ on$_{\!}$ visual$_{\!}$ recognition.$_{\!}$ With$_{\!}$ these$_{\!}$~ad- vancements,$_{\!}$ parametric$_{\!}$ softmax classifiers,$_{\!}$ which learn a set of parameters, \ie, weight vector, and bias term, for each class, have become the$_{\!}$ \textit{de}$_{\!}$ \textit{facto}$_{\!}$ regime$_{\!}$ in the area (Fig.$_{\!}$~\ref{fig:idea}(b)).$_{\!}$ However,$_{\!}$
due$_{\!}$ to$_{\!}$ the$_{\!}$ parametric$_{\!}$ nature,$_{\!}$~they suffer from several limitations: \textbf{First}, they lack simplicity and explainability. The parameters in the classification layer are abstract and detached from the physical nature of the problem being modelled$_{\!}$~\cite{angelov2020towards}.~Thus these classifiers are hard to naturally lend to an explanation that humans$_{\!}$ are$_{\!}$
able$_{\!}$ to$_{\!}$ process$_{\!}$~\cite{li2018deep}.$_{\!}$ \textbf{Second},$_{\!}$ linear$_{\!}$ classifiers$_{\!}$ are$_{\!}$ typically$_{\!}$ trained$_{\!}$ to$_{\!}$ optimize$_{\!}$ classification accuracy$_{\!}$ only,$_{\!}$ paying$_{\!}$ less$_{\!}$ attention$_{\!}$ to$_{\!}$ modeling$_{\!}$ the$_{\!}$ latent$_{\!}$ data$_{\!}$ structure.$_{\!}$ For$_{\!}$ each$_{\!}$ class,$_{\!}$ only$_{\!}$ one$_{\!}$ sin- gle$_{\!}$ weight$_{\!}$ vector$_{\!}$ is$_{\!}$ learned$_{\!}$ in$_{\!}$ a$_{\!}$ fully$_{\!}$ parametric$_{\!}$ manner.$_{\!}$ Thus$_{\!}$ they$_{\!}$ essentially$_{\!}$ assume$_{\!}$ \textit{unimodality}$_{\!}$ for
each~class$_{\!}$~\cite{hayashi2020discriminative,zhou2022rethinking},$_{\!}$ less$_{\!}$ tolerant$_{\!}$ of$_{\!}$ intra-class$_{\!}$ variation.$_{\!}$ 
\textbf{Third},$_{\!}$ as$_{\!}$ each$_{\!}$ class$_{\!}$ has$_{\!}$ its$_{\!}$ own$_{\!}$ set$_{\!}$ of$_{\!}$ parame-
ters,~deep parametric classifiers require the output space with a fixed dimensionality (equal to the number$_{\!}$ of$_{\!}$ classes)$_{\!}$~\cite{mettes2019hyperspherical}.$_{\!}$ As$_{\!}$ a$_{\!}$ result,$_{\!}$ their$_{\!}$ transferability$_{\!}$ is$_{\!}$ limited;$_{\!}$ when$_{\!}$ using$_{\!}$ ImageNet-trained classifiers  to initialize$_{\!}$ segmentation$_{\!}$~networks (\ie, pixel classifiers), the last classification layer, whose parameters are valuable knowledge learnt from the image classification task, has to be thrown away.

In$_{\!}$ light$_{\!}$ of$_{\!}$ the$_{\!}$ foregoing$_{\!}$ discussions,$_{\!}$ we$_{\!}$ are$_{\!}$ motivated$_{\!}$ to$_{\!}$ present$_{\!}$ \underline{d}eep$_{\!}$ \underline{n}earest$_{\!}$ \underline{c}entroids$_{\!}$ (DNC),$_{\!}$ a$_{\!}$ pow-$_{\!}$ erful, nonparametric classification network (Fig.$_{\!}$~\ref{fig:idea}(d)). Nearest Centroids, which has historical roots dating back to the dawn
of artificial$_{\!}$ intelligence$_{\!}$~\cite{fix1952discriminatory,cover1967nearest,kolodner1992introduction,kohonen1995learning,chaudhuri1996new,webb2003statistical},$_{\!}$ is$_{\!}$ arguably$_{\!}$ the$_{\!}$ simplest$_{\!}$ classifier.$_{\!}$ Nearest$_{\!}$ Centroids$_{\!}$ operates on an intuitive principle: given a data sample, it is directly classified to the~class of$_{\!}$ training$_{\!}$ examples$_{\!}$ whose$_{\!}$ mean$_{\!}$ (centroid)$_{\!}$ is$_{\!}$ closest$_{\!}$ to$_{\!}$ it.$_{\!}$ Apart$_{\!}$ from$_{\!}$ its$_{\!}$ internal transparency, Nearest Centroids is a classical form of exemplar-based reasoning$_{\!}$ \cite{kolodner1992introduction,li2018deep}, which is fundamental to our most effective strategies for tactical decision-making$_{\!}$~\cite{kim2014bayesian} (Fig.$_{\!}$~\ref{fig:idea}(c)). Numerous past studies$_{\!}$~\cite{newell1972human,rosch1973natural,aamodt1994case} have shown that humans~learn to$_{\!}$ solve$_{\!}$ new$_{\!}$ problems$_{\!}$ by$_{\!}$ using$_{\!}$ past$_{\!}$ solutions$_{\!}$ of$_{\!}$ similar$_{\!}$ problems.$_{\!}$ Despite$_{\!}$ its$_{\!}$~conceptual simplicity, empirical$_{\!}$ evidence$_{\!}$ in$_{\!}$ cognitive$_{\!}$ science,$_{\!}$ and$_{\!}$ ever$_{\!}$ popularity \cite{slade1991case,sanchez1997use,gou2012local,mensink2013distance},$_{\!}$ Nearest$_{\!}$ Centroids and its utility in large datasets with~high-dimensional input spaces are widely
unknown or ignored by current community. Inheriting the intuitive power of Nearest Centroids, our DNC is able to$_{\!}$ serve$_{\!}$ as$_{\!}$ a$_{\!}$ \textit{strong$_{\!}$ yet$_{\!}$ interpretable$_{\!}$ backbone$_{\!}$} for$_{\!}$ large-scale$_{\!}$ visual$_{\!}$ recognition;$_{\!}$ it$_{\!}$ is$_{\!}$ fully$_{\!}$ aware$_{\!}$ of the aforementioned limitations of parametric counterparts while shows even better performance.



$_{\!}$Specifically,$_{\!}$ DNC$_{\!}$ summarizes$_{\!}$ each$_{\!}$ class$_{\!}$ into$_{\!}$ a$_{\!}$ set$_{\!}$ of$_{\!}$ sub-centroids$_{\!}$ (sub-cluster$_{\!}$ centers)$_{\!}$ by$_{\!}$ clustering$_{\!}$ of training data inside the same class, and assigns each~test sample to the class$_{\!}$ with$_{\!}$ the nearest$_{\!}$ sub-centroid.$_{\!}$
DNC$_{\!}$ is$_{\!}$ essentially$_{\!}$ an$_{\!}$ \textit{experience}-/\textit{distance}-based$_{\!}$ classifier$_{\!}$ --$_{\!}$ it$_{\!}$ merely$_{\!}$ relies$_{\!}$ on$_{\!}$ the$_{\!}$ proximity$_{\!}$ of$_{\!}$ test$_{\!}$ query$_{\!}$ to$_{\!}$ local$_{\!}$ means$_{\!}$ of$_{\!}$ training$_{\!}$ data$_{\!}$ (``quintessential''$_{\!}$ past$_{\!}$ observations)$_{\!}$ in$_{\!}$ the$_{\!}$ deep$_{\!}$~feature$_{\!}$ space.$_{\!}$ As$_{\!}$ such,$_{\!}$ DNC$_{\!}$ learns$_{\!}$ visual$_{\!}$ recognition$_{\!}$ by$_{\!}$ directly$_{\!}$ optimizing$_{\!}$~the$_{\!}$ representation,$_{\!}$ instead$_{\!}$~of$_{\!}$~deep$_{\!}$\\
\noindent parametric$_{\!}$ models$_{\!}$ needing$_{\!}$ an$_{\!}$ extra$_{\!}$ softmax$_{\!}$ classification$_{\!}$ layer$_{\!}$ after$_{\!}$ feature$_{\!}$ extraction.$_{\!}$ For$_{\!}$ training,$_{\!}$ DNC$_{\!}$ alternates$_{\!}$ between$_{\!}$ two$_{\!}$ steps:$_{\!}$ \textbf{i)}$_{\!}$ \textit{class-wise}$_{\!}$ \textit{clustering}$_{\!}$ for$_{\!}$ automatically$_{\!}$ discovering$_{\!}$ class$_{\!}$ sub-centroids,$_{\!}$ and \textbf{ii)} \textit{classification prediction} for supervised representation learning, through retrieving\\
\noindent  the nearest$_{\!}$ sub-centroids.$_{\!}$ However,$_{\!}$ since$_{\!}$ the$_{\!}$ feature$_{\!}$ space$_{\!}$ evolves$_{\!}$ continually$_{\!}$~during training, computing the sub-centroids is expensive$_{\!}$ --$_{\!}$ it$_{\!}$ requires$_{\!}$ a$_{\!}$ pass over the$_{\!}$~full$_{\!}$ training$_{\!}$  dataset$_{\!}$ after$_{\!}$ each$_{\!}$ batch$_{\!}$ update and limits$_{\!}$ DNC's$_{\!}$ scalability.$_{\!}$ To$_{\!}$ solve$_{\!}$ this,$_{\!}$ we$_{\!}$ use a$_{\!}$ Sinkhorn$_{\!}$ Iteration$_{\!\!}$~\cite{cuturi2013sinkhorn}$_{\!}$ based$_{\!}$ clustering$_{\!}$ algorithm$_{\!}$~\cite{asano2020self}$_{\!}$ for$_{\!}$ fast$_{\!}$ cluster$_{\!}$ assignment.$_{\!}$ We$_{\!}$ further$_{\!}$ adopt$_{\!}$ momentum$_{\!}$ update$_{\!}$ with$_{\!}$ an$_{\!}$ external$_{\!}$ me- mory$_{\!}$ for$_{\!}$ estimating online the sub-centroids (whose amount$_{\!}$ is$_{\!}$ more$_{\!}$ than$_{\!}$ 1K$_{\!}$ on$_{\!}$ ImageNet$_{\!}$~\cite{ImageNet}) with small-batch size (\eg, 256). Consequently, DNC can be efficiently trained by simultaneously conducting clustering and stochastic optimization on large datasets with small batches, only slowing the training speed slightly (\eg, $\sim$5\% on ImageNet).


DNC$_{\!}$ enjoys$_{\!}$ a$_{\!}$ few$_{\!}$ attractive$_{\!}$ qualities:$_{\!}$ \textbf{First},$_{\!}$ improved$_{\!}$ \textbf{\textit{simplicity}}$_{\!}$ and$_{\!}$ \textbf{\textit{transparency}}.$_{\!}$ The$_{\!}$ intuitive$_{\!}$ working$_{\!}$ mechanism$_{\!}$ and$_{\!}$ statistical$_{\!}$ meaning$_{\!}$ of$_{\!}$ class$_{\!}$ sub-centroids$_{\!}$ make DNC elegant and easy to understand.  \textbf{Second}, automated discovery$_{\!}$ of$_{\!}$ \textbf{\textit{underlying$_{\!}$ data$_{\!}$ structure}}.$_{\!}$ By$_{\!}$ within-class$_{\!}$ deterministic$_{\!}$ clustering,$_{\!}$~the latent$_{\!}$ distribution$_{\!}$ of$_{\!}$ each$_{\!}$ class$_{\!}$ is$_{\!}$ automatically$_{\!}$ mined$_{\!}$ and$_{\!}$ fully$_{\!}$ captured$_{\!}$ as$_{\!}$ a$_{\!}$~set$_{\!}$~of$_{\!}$ representative$_{\!}$ local$_{\!}$ means.$_{\!}$ In$_{\!}$ contrast,$_{\!}$ parametric$_{\!}$ classifiers$_{\!}$ learn$_{\!}$ one$_{\!}$  single  weight$_{\!}$ vector$_{\!}$ per$_{\!}$ class,$_{\!}$ intolerant$_{\!}$ of$_{\!}$ rich$_{\!}$ intra-class$_{\!}$ variations.$_{\!}$ \textbf{Third}, direct supervision of \textbf{\textit{representation learning}}. DNC achieves classification by comparing
data samples and class sub-centroids on the feature space. With$_{\!}$ such$_{\!}$ distance-based$_{\!}$ nature,$_{\!}$ DNC$_{\!}$ blends$_{\!}$ unsupervised$_{\!}$ sub-pattern$_{\!}$ mining (class-wise$_{\!}$ clustering)$_{\!}$ and$_{\!}$ supervised$_{\!}$ representation$_{\!}$ learning$_{\!}$ (nonparametric$_{\!}$ classification)$_{\!}$ in$_{\!}$ a$_{\!}$ synergy:$_{\!}$ local$_{\!}$ significant$_{\!}$ pat- terns$_{\!}$ are$_{\!}$ automatically$_{\!}$ mined$_{\!}$~to$_{\!}$ facilitate$_{\!}$ classification$_{\!}$ decision-making;$_{\!}$ the$_{\!}$ supervisory$_{\!}$ signal$_{\!}$ from$_{\!}$
classification$_{\!}$ directly$_{\!}$ optimizes$_{\!}$ the$_{\!}$ representation,$_{\!}$ which$_{\!}$ in$_{\!}$ turn$_{\!}$ boosts$_{\!}$ meaningful$_{\!}$~clustering.$_{\!}$
\textbf{Forth},$_{\!}$ better$_{\!}$ \textit{\textbf{transferability}}.$_{\!}$ DNC$_{\!}$ learns$_{\!}$ by$_{\!}$~\textit{only}$_{\!}$~optimizing the feature representation, thus the output dimensionality no longer needs to be as many as the classes. With~this algorithmic merit, all~the~useful knowledge (parameters) learnt from a source~task (\eg,$_{\!}$ ImageNet$_{\!}$~\cite{ImageNet}$_{\!}$ classification)$_{\!}$ are$_{\!}$ stored$_{\!}$ in$_{\!}$ the$_{\!}$ representation$_{\!}$ space,$_{\!}$ and$_{\!}$ can$_{\!}$~be completely transferred to target tasks$_{\!}$ (\eg,$_{\!}$ Cityscapes$_{\!}$~\cite{cordts2016cityscapes} segmentation).$_{\!}$  \textbf{Fifth},$_{\!}$ \textit{\textbf{ad-hoc$_{\!\!}$ explainability}}.$_{\!}$ If$_{\!}$ further$_{\!}$ restricting$_{\!}$ the$_{\!}$ class$_{\!}$ sub-centroids$_{\!}$ to$_{\!}$ be$_{\!}$ samples (images)$_{\!}$ of$_{\!}$ the$_{\!}$ training$_{\!}$ set,$_{\!}$ DNC$_{\!}$ can$_{\!}$ explain$_{\!}$ its$_{\!}$ prediction$_{\!}$ based$_{\!}$ on$_{\!}$ \textit{IF}$_{\!\!}$ $\cdots$ $_{\!\!\!}$\textit{Then}$_{\!}$ rules$_{\!}$ and$_{\!}$ allow$_{\!}$ users to intuitively view the class representatives,$_{\!}$ and$_{\!}$ appreciate$_{\!}$ the$_{\!}$ similarity of test data$_{\!}$ to the representative images (detailed$_{\!}$ in$_{\!}$ \S\ref{sec:method}\&\ref{sec:adex}).$_{\!}$ Such$_{\!}$ \textit{ad-hoc}$_{\!}$ explainability$_{\!}$~\cite{biehl2016prototype} is valuable in safety-sensitive scenarios, and differs DNC from most existing network interpretation techniques$_{\!}$~\cite{simonyan2013deep,zeiler2014visualizing,cam2016learning} that only investigate$_{\!}$ \textit{post-hoc}$_{\!}$ explanations$_{\!}$ and$_{\!}$ thus$_{\!}$ fail$_{\!}$ to$_{\!}$ elucidate$_{\!}$ precisely$_{\!}$ how$_{\!}$ a$_{\!}$ model$_{\!}$ works$_{\!}$~\cite{lipton2018mythos,rudin2019stop}.


DNC is an intuitive yet general classification framework; it is compatible~with different visual~recognition network architectures and tasks.$_{\!}$ We$_{\!}$ experimentally$_{\!}$ show:$_{\!}$ \textbf{In$_{\!}$ \S\ref{sec:cls}},$_{\!}$ with$_{\!}$ ResNet$_{\!}$~\cite{he2016deep}$_{\!}$~and$_{\!}$ Swin$_{\!}$~\cite{liu2021swin}$_{\!}$\\
\noindent network$_{\!}$ architectures,$_{\!}$ DNC$_{\!}$~outperforms parametric counterparts on image classification, \ie,~\textbf{0.23-0.24}\%$_{\!}$ \texttt{top-1}$_{\!}$ accuracy$_{\!}$ on$_{\!}$ CIFAR-10$_{\!}$~\cite{CIFAR}$_{\!}$ and$_{\!}$ \textbf{0.24-0.32}\%$_{\!}$ on$_{\!}$ ImageNet$_{\!}$~\cite{ImageNet},$_{\!}$ by$_{\!}$ \textit{training$_{\!}$ from$_{\!}$ scratch}.\\
\noindent \textbf{In \S\ref{sec:seg}}, when using our ImageNet-pretrained, nonparametric versions of ResNet and Swin as back-\\
\noindent bones,$_{\!}$ our$_{\!}$ pixel-wise$_{\!}$ DNC$_{\!}$ classifier greatly improves the segmentation performance of FCN$_{\!}$~\cite{long2015fully}, DeepLab$_{\text{V3}\!\!}$~\cite{chen2017rethinking}, and UperNet$_{\!}$~\cite{xiao2018unified}, on ADE20K$_{\!}$~\cite{zhou2017scene} (\textbf{1.6-2.5}\% \texttt{mIoU}) and Cityscapes$_{\!}$~\cite{cordts2016cityscapes} (\textbf{1.1-1.9}\%$_{\!}$ \texttt{mIoU}).$_{\!}$ These$_{\!}$ results$_{\!}$ verify$_{\!}$ DNC's$_{\!}$ strong$_{\!}$ transferability$_{\!}$ and$_{\!}$ high$_{\!}$ versatility.$_{\!}$ \textbf{In$_{\!}$ \S\ref{sec:adex}},$_{\!}$ after$_{\!}$~con- straining class sub-centroids as training images of ImageNet, DNC becomes more interpretable, with only \textbf{0.12}\% sacrifice in \texttt{top-1} accuracy (but is still \textbf{0.17}\% better than the parametric counterpart).

These$_{\!}$ results$_{\!}$ are$_{\!}$ particularly$_{\!}$ impressive,$_{\!}$ considering$_{\!}$ the$_{\!}$ nonparametric$_{\!}$ and$_{\!}$ transparent$_{\!}$ nature$_{\!}$ of$_{\!}$ DNC.$_{\!}$ We feel this work brings fundamental insights into related fields. 

\vspace{-12pt}
\section{Related Work}\label{sec:rw}
\vspace{-10pt}
\noindent\textbf{Distance-/Prototype-based$_{\!}$ Classifiers.$_{\!\!}$} Among$_{\!}$ the$_{\!}$ numerous$_{\!}$ classification$_{\!}$ algorithms$_{\!}$ (\eg,$_{\!}$ logistic regression$_{\!}$~\cite{friedman2009elements},$_{\!}$ Na\"{i}ve$_{\!}$ Bayes$_{\!}$~\cite{hand2001idiot},$_{\!}$ random$_{\!}$ forest$_{\!\!}$~\cite{breiman2001random},$_{\!}$~support$_{\!}$ vector$_{\!}$ machines$_{\!}$~\cite{cortes1995support},$_{\!}$ and$_{\!}$ deep neural networks (DNNs)$_{\!}$~\cite{lecun2015deep}),$_{\!}$ distance-based$_{\!}$ methods are particularly remarkable, due to their intuitive working mechanism.$_{\!}$ Distance-based$_{\!}$ classifiers$_{\!}$ are$_{\!}$ nonparametric$_{\!}$ and$_{\!}$ exemplar-driven,$_{\!}$ relying$_{\!}$ on$_{\!}$~similarities between samples and internally stored exemplars/prototypes. Thus they conduct~case-based reasoning that humans use naturally in problem-solving, making them appealing and interpretable$_{\!}$~\cite{newell1972human,rudin2022interpretable}. $k$-Nearest$_{\!}$ Neighbors$_{\!}$ ($k$-NN)$_{\!}$~\cite{fix1952discriminatory,cover1967nearest}$_{\!}$ is a form of distance-based classifiers; it~uses \textit{all} training data as exemplars$_{\!}$~\cite{boiman2008defense,guillaumin2009tagprop}. Towards network implementation of $k$-NN$_{\!}$~\cite{lippmann1989pattern,salakhutdinov2007learning,papernot2018deep},$_{\!}$ Wu$_{\!}$ \etal$_{\!}$~\cite{wu2018improving} made notable progress; their $k$-NN network outperforms parametric softmax based ResNet$_{\!\!}$~\cite{he2016deep} and the learnt representation works well in few-shot settings. However, $k$-NN classifiers (including the deep learning analogues) cost huge storage space and pose heavy computation burden (\eg, persistently retaining the training dataset and making full-dataset retrieval for each query)$_{\!}$~\cite{garcia2012prototype,yang2018robust}, and the nearest neighbors may not be good class representatives$_{\!}$~\cite{rudin2022interpretable}. Nearest Centroids$_{\!}$~\cite{kolodner1992introduction,kohonen1995learning,chaudhuri1996new,webb2003statistical} is another famous distance-based classifier yet has neither of the deficiencies of $k$-NN$_{\!}$~\cite{sanchez1997use,rudin2022interpretable}. Nearest Centroids selects representative$_{\!}$ class$_{\!}$ centers,$_{\!}$ instead$_{\!}$ of$_{\!}$ all$_{\!}$ the$_{\!}$ training$_{\!}$ data,$_{\!}$ as$_{\!}$ exemplars.$_{\!}$ Guerriero$_{\!}$ \etal$_{\!}$~\cite{guerriero2018deepncm} also investigate the idea of bringing Nearest Centroids into DNNs. However, they simply abstract each class into one single class mean, failing to capture complex class-wise distributions and showing weak results even in small datasets$_{\!}$~\cite{CIFAR}.

\vspace{-3pt}
The idea of distance-based classification also stimulates the emergence of \textit{prototypical} \textit{networks}, which mainly focus on few-shot$_{\!}$~\cite{snell2017prototypical,dong2018few} and zero-shot$_{\!}$~\cite{jetley2015prototypical,xu2020attribute}$_{\!}$ learning. However, they often associate to each class only one representation (prototype)$_{\!}$~\cite{rebuffi2017icarl} and their prototypes are usually  flexible parameters$_{\!}$~\cite{snell2017prototypical,yang2018robust,xu2020attribute} or defined prior to training$_{\!}$~\cite{jetley2015prototypical,mettes2019hyperspherical}. In DNC, a prototype (sub-centroid)~is either a generalization of a number of observations or intuitively a typical training visual example. Via clustering based sub-class mining, DNC addresses two key properties of prototypical exemplars: \textit{sparsity} and \textit{expressivity}$_{\!}$~\cite{quattoni2008transfer,biehl2009metric}. In this way, the representation can be learnt to capture the underlying class structure, hence facilitating large-scale visual recognition while preserving transparency. 

\vspace{-3pt}
\noindent\textbf{Neural$_{\!}$ Network$_{\!}$ Interpretability.$_{\!}$} As$_{\!}$ the$_{\!}$ \textit{black-box}$_{\!}$ nature limits the adoption of DNNs in decision-critical tasks, there has been a recent surge of interest in DNNs' interpretability. However, most interpretation techniques only produce posteriori explanations for already-trained DNNs, typically$_{\!}$ by$_{\!}$ analysis$_{\!}$ of$_{\!}$ reverse-engineer$_{\!}$ importance$_{\!}$ values$_{\!}$~\cite{erhan2009visualizing,simonyan2013deep,zeiler2014visualizing,mahendran2015understanding,yosinski2015understanding,bach2015pixel,cam2016learning,shrikumar2017learning,selvaraju2017grad}$_{\!}$ and$_{\!}$  sensitivities$_{\!}$ of$_{\!}$ inputs$_{\!}$~\cite{ribeiro2016should,zintgraf2017visualizing,koh2017understanding,lundberg2017unified}. As many literature outlined, \textit{post-hoc} explanations are problematic and misleading~$_{\!}$\cite{laugel2019dangers,rudin2019stop,arrieta2020explainable,rudin2022interpretable};$_{\!}$ they$_{\!}$ cannot$_{\!}$ explain$_{\!}$ what$_{\!}$ \textit{actually}$_{\!}$ makes$_{\!}$ a$_{\!}$ DNN$_{\!}$ arrive$_{\!}$ at$_{\!}$ its$_{\!}$ decisions$_{\!}$ \cite{chen2019looks}.$_{\!}$ To$_{\!}$ pursue$_{\!}$ \textit{ad-hoc}$_{\!}$ explainability, some  attempts have been initiated to develop explainable DNNs, by deploying more interpretable machineries into black-box DNNs$_{\!}$~\cite{alvarez2018towards,kim2018interpretability,wang2019gaining} or regularizing representation with certain properties (\eg, sparsity $_{\!}$\cite{subramanian2018spine}, decomposability$_{\!}$~\cite{chen2016infogan}, monotonicity$_{\!}$~\cite{you2017deep}) that can enhance interpretability.

\vspace{-3pt}
DNC intrinsically relies on class sub-centroid retrieving. The theoretical simplicity makes it easy to understand; when anchoring the sub-centroids to available observations, DNC can derive intuitive explanations based on the similarities of test samples to representative observations. It simultaneously conducts representation learning and case-based reasoning, making it self-explainable without post-hoc analysis$_{\!}$~\cite{biehl2016prototype}. DNC relates to \textit{concept}-based explainable networks$_{\!}$~\cite{alvarez2018towards,li2018deep,kim2018interpretability,ming2019interpretable,chen2019looks,arik2020protoattend,angelov2020towards,nauta2021neural} that\\
\noindent refer to human-friendly concepts/prototypes during decision making. These methods, however, necessitate nontrivial architectural modification and usually resort to pre-trained models, not to men- tion$_{\!}$ serving$_{\!}$ as$_{\!}$~backbone$_{\!}$ networks.$_{\!}$ In$_{\!}$ sharp contrast,$_{\!}$ DNC$_{\!}$ only$_{\!}$ brings$_{\!}$ minimal$_{\!}$ architectural$_{\!}$ change to~parametric classifier based DNNs and yields remarkable performance on ImageNet$_{\!}$~\cite{ImageNet} with training from scratch and  \textit{ad-hoc} explainability. It provides solid empirical evidence, for the first time as far as we know, for the power of case-based reasoning in large-scale visual recognition.







\vspace{-7pt}
\section{Deep Nearest Centroids (DNC)}
\label{sec:method}
\vspace{-5pt}
\noindent\textbf{$_{\!}$Problem$_{\!}$ Statement.$_{\!}$} Consider$_{\!}$ the$_{\!}$ standard$_{\!}$ visual$_{\!}$ recognition$_{\!}$ setting.$_{\!}$ Let$_{\!}$ $\mathcal{X}$$_{\!}$~be$_{\!}$ the$_{\!}$ visual$_{\!}$ space$_{\!}$ (\eg,$_{\!}$ image$_{\!}$ space$_{\!}$ for$_{\!}$ recognition,$_{\!}$ pixel$_{\!}$ space$_{\!}$ for$_{\!}$ segmentation),$_{\!}$ and$_{\!}$ $\mathcal{Y}\!=\!\{1,_{\!} \cdots_{\!}, C\}_{\!}$ the$_{\!}$ set$_{\!}$ of$_{\!}$ semantic$_{\!}$ classes.$_{\!}$ Given$_{\!}$ a$_{\!}$ training dataset$_{\!}$ $\{(x_n,y_n)_{\!}\!\in_{\!}\!\mathcal{X}_{\!}\!\times_{\!}\!\mathcal{Y}\}_{n=1}^N$,$_{\!}$ the$_{\!}$ goal$_{\!}$ is$_{\!}$ to$_{\!}$ use$_{\!}$ the$_{\!}$ $N_{\!}$ training$_{\!}$ examples$_{\!}$ to$_{\!}$ fit$_{\!}$ a$_{\!}$ \textit{model}$_{\!}$ (or$_{\!}$ hypothesis)$_{\!}$ $h_{\!\!}:_{\!\!\!\!}~\mathcal{X}\!\mapsto\!\mathcal{Y}$$_{\!}$ that$_{\!}$ accurately$_{\!}$ predicts$_{\!}$ the$_{\!}$ semantic$_{\!}$ classes$_{\!}$ for$_{\!}$ new$_{\!}$ visual$_{\!}$ samples.$_{\!}$

\noindent\textbf{$_{\!}$Parametric$_{\!}$ Softmax$_{\!}$ Classifier.$_{\!}$} Current$_{\!}$ common$_{\!}$ practice$_{\!}$ is$_{\!}$ to$_{\!}$ implement$_{\!}$ $h$$_{\!}$~as DNNs$_{\!}$ and$_{\!}$ decompose$_{\!}$ it$_{\!}$ as$_{\!}$ $h\!=\!l_{\!}\circ_{\!}\!f$.$_{\!}$ Here$_{\!}$ $f_{\!}:_{\!\!}~\mathcal{X}\!\mapsto\!\mathcal{F}_{\!}$ is$_{\!}$ a$_{\!}$ \textit{feature}$_{\!}$ \textit{extractor}$_{\!}$ (\eg,$_{\!}$~convolution$_{\!}$ based$_{\!}$ or$_{\!}$ Transformer-like$_{\!}$ networks)$_{\!}$ that$_{\!}$ maps$_{\!}$ an$_{\!}$ input$_{\!}$ sample$_{\!}$ $x_{\!\!}\in_{\!\!}\mathcal{X}$~into a $d$-dimensional representation space$_{\!}$ $\mathcal{F}\!\in\!\mathbb{R}^d$, \ie, $\bm{x}\!=\!f(x)\!\in\!\mathcal{F}$; and $l_{\!\!}:_{\!\!\!\!}~\mathcal{F}\!\mapsto\!\mathcal{Y}_{\!}$ is~a \textit{parametric classifier} (\eg, the last fully-connected layer in recognition or last~1$\times$1 convolution$_{\!}$ layer$_{\!}$ in$_{\!}$ segmentation)$_{\!}$ that$_{\!}$~takes$_{\!}$~$\bm{x}$$_{\!}$ as$_{\!}$ input$_{\!}$ and$_{\!}$ produces$_{\!}$ class$_{\!}$ prediction$_{\!}$ $\hat{y}_{\!}=_{\!}l(\bm{x})\!\in\!\mathcal{Y}$.$_{\!}$ Concretely,
$l$ assigns a query $x\!\in\!\mathcal{X}$ to the class $\hat{y}\!\in\!\mathcal{Y}$ according to:
\vspace{-3pt}
\begin{equation}\label{eq:psoftmax1}
\small\setlength\belowdisplayskip{2pt}
\hat{y}\!=\!\mathop{\arg\max}\nolimits_{c\!\!~\in\!\!~\mathcal{Y}} s^c,~~~~~s^c\!=\!(\bm{w}^c)^{\!\top}\!\bm{x}\!+\!b^c,
\end{equation}
where $s^c\!\in\!\mathbb{R}$ indicates the unnormalized  prediction score (\ie, \textit{logit}) for class $c$,  $\bm{w}^c\!\in\!\mathbb{R}^d$ and $b^c\!\in\!\mathbb{R}$ are learnable parameters -- class weight and bias term for $c$. Parameters of $l$ and $f$ are learnt~by~mini- mizing the softmax cross-entropy loss:
\vspace{-5pt}
\begin{equation}\label{eq:psoftmax2}
\small\setlength\belowdisplayskip{2pt}
\mathcal{L} \!=\!\frac{1}{N}\sum^N\nolimits_{n=1}\!-\log p(y_n|x_n),~~~~ p(y|x) \!=\!\mathrm{softmax}_y(l_{\!}\circ_{\!}\!f(x)) \!=\! \frac{\exp(s^y)}{\sum_{c\in\mathcal{Y}}\exp(s^c)}.
\end{equation}
Though$_{\!}$ highly$_{\!}$ successful,$_{\!}$ the$_{\!}$ use$_{\!}$ of$_{\!}$ the$_{\!}$ parametric$_{\!}$ classifier$_{\!}$ $l$$_{\!}$ has$_{\!}$ drawbacks$_{\!}$~as well: \textbf{i)}$_{\!}$ The$_{\!}$ weight$_{\!}$ matrix$_{\!}$ $\bm{W}\!\!=\!(\bm{w}^1,_{\!}\cdots_{\!},\bm{w}^C)\!\in\!\mathbb{R}^{d\times C\!\!}$ and$_{\!}$ bias$_{\!}$ vector$_{\!}$ $\bm{b}\!=\!({b}^1,_{\!}\cdots_{\!}, {b}^C)$ $\!\in\!\mathbb{R}^d$ in $l$ are learnable parameters, which cannot provide any information about what makes the model $h$ reach its decisions. \textbf{ii)} $l$ makes
the$_{\!}$ loss$_{\!}$ $\mathcal{L}_{\!}$ only$_{\!}$ depend$_{\!}$ on$_{\!}$ the$_{\!}$ relative$_{\!}$ relation$_{\!}$ among$_{\!}$ logits,$_{\!}$ \ie,$_{\!}$ $\{s^c\}_{c}$,$_{\!}$ and$_{\!}$ cannot$_{\!}$ directly$_{\!}$ supervise$_{\!}$ on the$_{\!}$ representation$_{\!}$ $\bm{x}$$_{\!}$~\cite{pang2020rethinking,zhang2020rbf}.$_{\!}$ \textbf{iii)}$_{\!}$ $\bm{W}_{\!}$ and$_{\!}$ $\bm{b}_{\!}$ are$_{\!}$ learnt$_{\!}$ as$_{\!}$ flexible$_{\!}$ parameters,$_{\!}$ lacking$_{\!}$ explicit$_{\!}$ modeling$_{\!}$ of$_{\!}$ the$_{\!}$ underlying$_{\!}$ data$_{\!}$ structure.$_{\!}$ \textbf{iv)}$_{\!}$ The$_{\!}$ final$_{\!}$ output$_{\!}$ dimensionality$_{\!}$ is$_{\!}$ {constrained$_{\!}$ to$_{\!}$ be}$_{\!}$ the$_{\!}$ number$_{\!}$ of classes, \ie, $C$. During transfer learning, as different visual recognition tasks typically have distinct semantic$_{\!}$ label$_{\!}$ spaces$_{\!}$ (with$_{\!}$ different$_{\!}$ number$_{\!}$~of$_{\!}$ classes),$_{\!}$ the$_{\!}$ classifier$_{\!}$ from$_{\!}$ a$_{\!}$ pretrained$_{\!}$ model$_{\!}$ has$_{\!}$ to$_{\!}$ be abandoned,$_{\!}$ even$_{\!}$ though$_{\!}$ the$_{\!}$  learnt$_{\!}$ parameters$_{\!}$ $\bm{W}_{\!}$ and$_{\!}$ $\bm{b}_{\!}$ are$_{\!}$ valuable$_{\!}$ knowledge$_{\!}$ from$_{\!}$ the$_{\!}$ source task.

The question naturally arises:  might there be a simple way to address these limitations of current~\textit{de} \textit{facto}, parametric classifier based visual recognition regime? Here we show that this is indeed possible, even with better performance.


\begin{figure*}[t]
\vspace{-12pt}
  \centering
      \includegraphics[width=1 \linewidth]{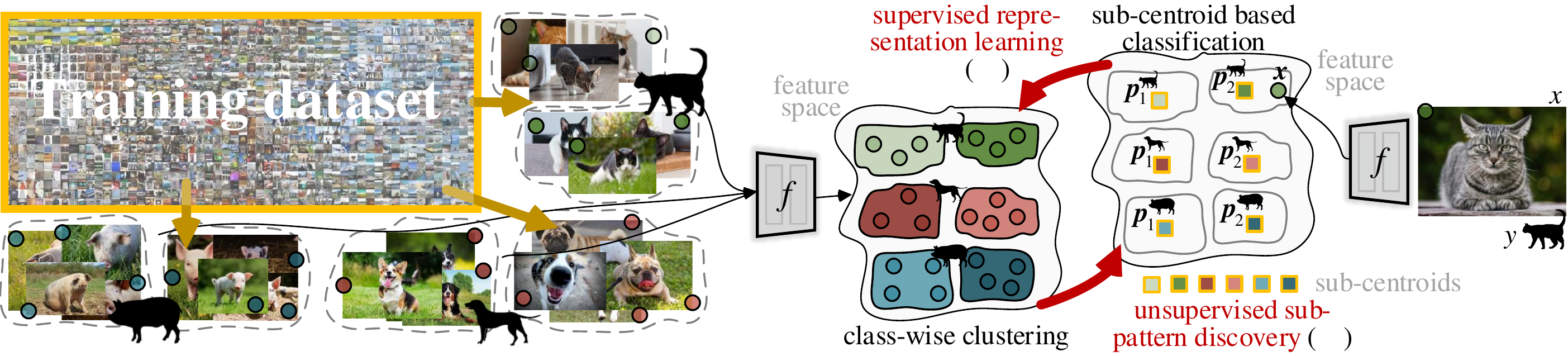}
     \put(-62,1.7){\scriptsize{\ref{eq:BIP1}}}
     \put(-148.7,70){\scriptsize{\ref{eq:DNC3}}}
     \put(-159,70){\scriptsize{$\mathcal{L}$}}
     \vspace{-7pt}
\caption{$_{\!}$With$_{\!}$ a$_{\!}$ distance-/case-based$_{\!}$ classification$_{\!}$ scheme,$_{\!}$ DNC$_{\!}$ combines$_{\!}$ unsupervised sub-pattern discovery and supervised representation learning in a synergy. }
\label{fig:framework}
\vspace{-10pt}
\end{figure*}

\noindent\textbf{DNC Classifier.} Our DNC (Fig.$_{\!}$~\ref{fig:framework}) is built upon the intuitive idea of Nearest Centroids, \ie, assign a sample $x$ to the class $\hat{y}\!\in\!\mathcal{Y}$ with the closest class center:
\vspace{-3pt}
\begin{equation}\label{eq:DNC1}
\small\setlength\belowdisplayskip{2pt}
\hat{y}\!=\!\mathop{\arg\min}\nolimits_{c\!\!~\in\!\!~\mathcal{Y}} \langle\bm{x}, \bar{\bm{x}}^c\rangle,~~~~\bar{\bm{x}}^c\!=\!\frac{1}{N^c}\sum\nolimits_{{x}^c_n:\!~y^c_n=c}\bm{x}^c_n,
\end{equation}
where $\langle\cdot,\cdot\rangle$ is a distance measure, given as: $\langle\bm{u}, \bm{v}\rangle\!=\!-\bm{u}^{\!\!\top\!\!}\bm{v}/\|\bm{u}\|\|\bm{v}\|$.$_{\!}$  For simplicity,  all the features are$_{\!}$ defaulted$_{\!}$ to$_{\!}$ $\ell_2$-normalized$_{\!}$ from$_{\!}$ now$_{\!}$ on.$_{\!}$ $\bar{\bm{x}}^{c\!}$ is$_{\!}$ the$_{\!}$ mean$_{\!}$ vector$_{\!}$ of$_{\!}$ class$_{\!}$ $c$,$_{\!}$ ${x}^c_n$$_{\!}$ is$_{\!}$ a$_{\!}$ training$_{\!}$  sample$_{\!}$ of$_{\!}$ $c$,$_{\!}$~\ie,$_{\!}$ $y^c_n\!=\!c$,$_{\!}$ and$_{\!}$ $N^{c\!}$ is$_{\!}$ the$_{\!}$ number$_{\!}$ of$_{\!}$ training$_{\!}$ samples$_{\!}$ in$_{\!}$ $c$.$_{\!}$ As$_{\!}$ such,$_{\!}$~the feature-to-class$_{\!}$ mapping$_{\!}$ $\mathcal{F}\!\mapsto\!\mathcal{Y}$$_{\!}$ is$_{\!}$ achieved$_{\!}$ in$_{\!}$ a$_{\!}$ \textit{nonparametric}$_{\!}$ manner$_{\!}$ and$_{\!}$ \textit{understandable}$_{\!}$  from user's$_{\!}$ view,$_{\!}$ in$_{\!}$ contrast$_{\!}$ to$_{\!}$ the$_{\!}$ parametric$_{\!}$ classifier$_{\!}$ $l$$_{\!}$ that$_{\!}$ learns$_{\!}$~``non-transparent'' parameters$_{\!}$ for$_{\!}$ each$_{\!}$ class.$_{\!}$ It$_{\!}$ makes$_{\!}$ more$_{\!}$ sense if$_{\!}$~multiple$_{\!}$ sub-centroids$_{\!\!}$ (local$_{\!}$ means) per class are used, which is in particular true for challenging visual recognition  where complex intra-class variations cannot be simply$_{\!}$ described$_{\!}$ by$_{\!}$ the$_{\!}$ simple assumption of unimodality of data of each class.$_{\!\!}$ When$_{\!}$ representing$_{\!}$ each$_{\!}$ class$_{\!}$ $c$$_{\!}$ as$_{\!}$~$K$$_{\!}$~sub-centroids,$_{\!}$ denoted$_{\!}$ by$_{\!}$ $\{\bm{p}^c_{k\!}\!\in\!\mathbb{R}^d\}_{k=1}^K$,$_{\!}$ the$_{\!}$ $C$-way$_{\!}$ classification$_{\!}$ for$_{\!}$ sample$_{\!}$ $x_{\!}$ takes$_{\!}$ place$_{\!}$ as$_{\!}$   a$_{\!}$ \textit{winner-takes-all}$_{\!}$ rule:
\vspace{-5pt}
\begin{equation}\label{eq:DNC2}
\small\setlength\belowdisplayskip{2pt}
\hat{y}\!=\!c^*, ~~~~(c^*, k^*)=\mathop{\arg\min}\nolimits_{\!~c\!~\in\!~\mathcal{Y},  k\!~\in\!~\{1,\cdots,K\}} \langle\bm{x},{\bm{p}}^c_k\rangle.
\end{equation}
Clearly,$_{\!}$ estimating$_{\!}$ class$_{\!}$ sub-centroids$_{\!}$ needs$_{\!}$ clustering$_{\!}$ of$_{\!}$ training$_{\!}$ samples$_{\!}$ within  each$_{\!}$ class.$_{\!}$ As$_{\!}$ class

sub-centroids$_{\!}$ are$_{\!}$ sub-cluster$_{\!}$ centers$_{\!}$ in$_{\!}$ the$_{\!}$ latent$_{\!}$ feature$_{\!}$ space$_{\!}$ $\mathcal{F}$, they are locally significant visual patterns and can comprehensively represent class-level characteristics. DNC can  be intuitively understood as selecting and storing prototypical exemplars for each class, and finding classification evidence for$_{\!}$ a$_{\!}$ previously$_{\!}$ unseen$_{\!}$ sample$_{\!}$ by$_{\!}$ retrieving$_{\!}$ the$_{\!}$ most$_{\!}$ similar$_{\!}$ exemplar.$_{\!}$ This$_{\!}$ also aligns$_{\!}$  with$_{\!}$  the$_{\!}$  \textit{prototype}$_{\!}$  \textit{theory}$_{\!}$  in$_{\!}$ psychology$_{\!}$~\cite{rosch1973natural,rosch1975cognitive,knowlton1993learning}:$_{\!}$  prototypes$_{\!}$  are$_{\!}$  a$_{\!}$  typical form of cognitive organisation of real world objects. DNC thus emulates the case-based reasoning process that we humans are accustomed to$_{\!}$~\cite{biehl2016prototype}. For instance, when ornithologists classify a bird, they will compare it with those typical exemplars from known bird species to decide which species the bird belongs to$_{\!}$~\cite{rudin2022interpretable}.






\noindent\textbf{Sub-centroid Estimation.} To find informative sub-centroids that best represent classes, we perform deterministic clustering within each class on the representation space $\mathcal{F}$. More specifically, for each class $c$, we cluster all the representations $\{\bm{x}^c_{n\!}\!\in\!\mathbb{R}^d\}_{n=1\!}^{N^c}$ into$_{\!}$ $K$ clusters whose centers are used as the sub-centroids of $c$, \ie, $\{\bm{p}^c_{k\!}\!\in\!\mathbb{R}^d\}_{k=1}^K$.$_{\!}$ Let$_{\!}$ $\bm{X}^{c\!}\!=\![\bm{x}^c_1, \cdots_{\!}, \bm{x}^c_{N^c}]\!\in\!\mathbb{R}^{d\times N^c\!\!\!}$ and$_{\!}$ $\bm{P}^{c\!}\!=\![\bm{p}^c_1, \cdots_{\!}, \bm{p}^c_{K}]\!\in\!\mathbb{R}^{d\times K\!}$ denote$_{\!}$ the$_{\!}$ feature$_{\!}$ and$_{\!}$ sub-centroid$_{\!}$ matrixes,$_{\!}$  respectively.$_{\!}$ The$_{\!}$ deterministic$_{\!}$ clustering, \ie, the mapping from $\bm{X}^{c\!}$ to $\bm{P}^{c\!}$, can be denoted as $\bm{Q}^{c\!}\!=\![\bm{q}^c_1, \cdots_{\!}, \bm{q}^c_{N^c}]\!\in\!\{0,1\}^{K\times N^c\!}$, where $n$-\textit{th} column $\bm{q}^c_n\!\in\!\{0,1\}^{K\!}$ is an one-hot assignment vector of $n$-\textit{th} sample~$x^c_n$ \textit{w.r.t} the $K$ clusters. $\bm{Q}^{c\!}$ is desired to maximize the similarity between $\bm{X}^{c\!}$ and $\bm{P}^{c\!}$, leading to the following binary integer program (BIP):
\begin{equation}
\small\setlength\belowdisplayskip{2pt}
\begin{aligned}\label{eq:BIP1}
\!\!\!\!\max_{\bm{Q}^{c}\in\mathcal{Q}^{c}}\!\texttt{Tr}\big((\bm{Q}^{c})^{\!\top\!}(\bm{P}^{c})^{\!\top\!}\bm{X}^c\big),
~~\!\mathcal{Q}^{c}\!=\!\{\bm{Q}^{c\!}\!\in\!\{0,1\}^{K\!\times\!N^c}|(\bm{Q}^{c})^{\!\top}\!\bm{1}_K\!=\!\bm{1}_{N^c}\},
\end{aligned}
\end{equation}
where$_{\!}$ $\bm{1}_{K\!}$ is$_{\!}$ a$_{\!}$ $K$-dimensional$_{\!}$ all-ones$_{\!}$ vector.$_{\!}$ As$_{\!}$ in$_{\!}$~\cite{asano2020self},$_{\!}$ we$_{\!}$ relax$_{\!}$ $\mathcal{Q}^{c\!}$ to$_{\!}$ be$_{\!}$ a$_{\!}$ \textit{transportation}$_{\!}$ \textit{polytope}$_{\!\!}$~\cite{cuturi2013sinkhorn}:$_{\!}$ $\mathcal{Q}'^{c\!}_{\!}\!=_{\!}\!\{\bm{Q}^{c\!}\!\in\!\mathbb{R}^{K\!\times\!N^c\!}_+|(\bm{Q}^{c})^{\!\top}\!\bm{1}_{\!K{\!}}\!=_{\!}\!\bm{1}_{\!N^c}, \bm{Q}^{c}\bm{1}_{\!N^c\!}\!=_{\!}\!\frac{N^c}{K}_{\!}\bm{1}_{\!K\!}\}$,$_{\!\!}$ casting$_{\!}$ BIP$_{\!\!}$~(\ref{eq:BIP1})$_{\!}$ into$_{\!}$ an$_{\!\!}$ \textit{optimal$_{\!}$ transport}$_{\!}$ problem.$_{\!}$ In$_{\!}$ $\mathcal{Q}'^{c\!}$,$_{\!}$ besides$_{\!}$ the$_{\!}$ \textit{one-hot}$_{\!}$ \textit{assignment} constraint (\ie, $(\bm{Q}^{c})^{\!\top}\!\bm{1}_{\!K{\!}}\!=_{\!}\!\bm{1}_{\!N^c}$),$_{\!}$ an$_{\!}$ \textit{equipartition}$_{\!}$ constraint$_{\!}$ (\ie, $\bm{Q}^{c}\bm{1}_{\!N^c\!}\!=_{\!}\!\frac{N^c}{K}_{\!}\bm{1}_{\!K\!}$) is$_{\!}$ added$_{\!}$ to$_{\!}$ inspire$_{\!}$ $N^c$$_{\!}$ samples$_{\!}$ to$_{\!}$ be$_{\!}$~evenly$_{\!}$ assigned$_{\!}$ to$_{\!}$ $K_{\!}$ clusters.$_{\!}$ This$_{\!}$~can$_{\!}$ efficiently$_{\!}$ avoid$_{\!}$  degeneracy,$_{\!}$ \ie,$_{\!}$ mapping$_{\!}$ all the$_{\!}$ data$_{\!}$ to$_{\!}$ the$_{\!}$ same$_{\!}$ cluster.$_{\!}$ Then$_{\!}$ the$_{\!}$ solution$_{\!}$ can$_{\!}$~be$_{\!}$ given$_{\!}$ by$_{\!}$ a$_{\!}$~fast$_{\!}$ version$_{\!}$~\cite{cuturi2013sinkhorn}$_{\!}$
of Sinkhorn-Knopp algorithm$_{\!}$~\cite{knight2008sinkhorn}, in a form of a normalized exponential matrix:
\vspace{-4pt}
\begin{equation}
\small
	\begin{aligned}\label{eq:BIP2}
		{\bm{Q}}^{c*}  = \mathrm{diag}(\bm{\alpha}) \exp \big(\frac{(\bm{P}^{c})^{\!\top\!}\bm{X}^c}{\varepsilon}\big)\mathrm{diag}(\bm{\beta}),
	\end{aligned}
\end{equation}
where the exponentiation is performed element-wise, $\bm{\alpha}\!\in\!\mathbb{R}^{K\!\!}$ and $\bm{\beta}\!\in\!\mathbb{R}^{N^c\!\!}$ are two renormalization vectors, which can be computed using a small number of matrix multiplications via Sinkhorn-Knopp Iteration$_{\!}$~\cite{cuturi2013sinkhorn}, and $\varepsilon\!=\!0.05$ trades off convergence$_{\!}$ speed$_{\!}$ with$_{\!}$ closeness$_{\!}$ to$_{\!}$ the$_{\!}$ original$_{\!}$ transport$_{\!}$ problem.$_{\!}$ In$_{\!}$ short,$_{\!}$ by$_{\!}$ mapping data$_{\!}$  samples$_{\!}$  into$_{\!}$  a$_{\!}$  few$_{\!}$  clusters$_{\!}$  under$_{\!}$  the$_{\!}$  constraints$_{\!}$  $\mathcal{Q}'^{c}$,$_{\!}$  we$_{\!}$  pursue$_{\!}$~\textit{sparsity}$_{\!}$  and$_{\!}$ \textit{expressivity}$_{\!}$~\cite{quattoni2008transfer,biehl2009metric},$_{\!}$ making$_{\!}$ class$_{\!}$ sub-centroids$_{\!}$ representative$_{\!}$ of$_{\!}$ the$_{\!}$ dataset.



\noindent\textbf{Training of DNC$_{\!}$ =$_{\!}$ Supervised$_{\!}$ Representation$_{\!}$ Learning$_{\!\!}$ +$_{\!\!}$ Automatic Sub-class$_{\!}$ Pattern$_{\!}$ Mining.$_{\!}$} Ideally, according to class-wise cluster assignments $\{{\bm{Q}}^{c\!}\}_{c=1}^C$,$_{\!}$ we$_{\!}$ can$_{\!}$ get$_{\!}$ totally$_{\!}$  $CK$$_{\!}$  sub-centroids$_{\!}$  $\{\bm{p}^c_k\}_{c,k=1}^{C,K}$,$_{\!}$  \ie,$_{\!}$  mean$_{\!}$  feature$_{\!}$  vectors of the training data in the $CK$ clusters. Then the training target becomes as:
\vspace{-8pt}
\begin{equation}\label{eq:DNC3}
\small
~~~~~\mathcal{L} \!=\!\frac{1}{N}\sum^N\nolimits_{n=1}\!-\log p(y_n|x_n),~~ p(y|x) \!=\!\frac{\exp\big(-\min (\{ \langle \bm{x}, \bm{p}^y_{k} \rangle \}_{k=1}^K)\big)}{\sum_{c\in\mathcal{Y}\!}\exp\big(-\min( \{ \langle \bm{x}, \bm{p}^c_{k} \rangle \}_{k=1}^K)\big)}.
\end{equation}
Comparing$_{\!}$ (\ref{eq:psoftmax2})$_{\!}$ and$_{\!}$ (\ref{eq:DNC3}),$_{\!}$ as$_{\!}$ the$_{\!}$ class$_{\!}$ sub-centroids$_{\!}$  $\{\bm{p}^c_k\}_{c,k\!}$ are derived solely from data representations, DNC learns visual recognition by directly optimizing the~representation $f$, instead of the parametric classifier $l$. Moreover, with such a nonparametric, distance-based scheme, DNC builds a closer link to metric learning$_{\!}$~\cite{hadsell2006dimensionality,sohn2016improved,kaya2019deep,boudiaf2020unifying,sun2020circle,suzuki2019arc,khosla2020supervised};$_{\!}$ DNC$_{\!}$ can$_{\!}$ even$_{\!}$ be$_{\!}$ viewed$_{\!}$ as$_{\!}$ learning$_{\!}$ a$_{\!}$ metric function$_{\!}$ $f$$_{\!}$ to$_{\!}$ compare$_{\!}$ data$_{\!}$ samples $\{x_{n\!}\}_{n}$, under the guidance of the corresponding semantic labels $\{y_{n\!}\}_n$.



$_{\!}$During$_{\!}$ training,$_{\!}$ DNC$_{\!}$ alternates$_{\!}$ two$_{\!}$ steps$_{\!}$ iteratively:$_{\!}$ \textbf{i)}$_{\!}$ class-wise$_{\!}$ clustering (\ref{eq:BIP1}) for$_{\!}$ automatic$_{\!}$ sub-centroid$_{\!}$ discovery,$_{\!}$ and$_{\!}$ \textbf{ii)}$_{\!}$ sub-centroid$_{\!}$ based$_{\!}$ classification for$_{\!}$ supervised$_{\!}$ representation$_{\!}$ learning$_{\!}$ (\ref{eq:DNC3}).$_{\!}$ Through$_{\!}$ clustering,$_{\!}$ DNC$_{\!}$ probes$_{\!}$ underlying$_{\!}$ data$_{\!}$ distribution$_{\!}$ of$_{\!}$ each$_{\!}$ class,$_{\!}$ and$_{\!}$ produces$_{\!}$ informative
sub-centroids$_{\!}$ by$_{\!}$ aggregating$_{\!}$ statistics$_{\!}$ from$_{\!}$ data$_{\!}$ clusters.$_{\!}$ This$_{\!}$ automatic$_{\!}$ sub-class$_{\!}$ discovery$_{\!}$ process also$_{\!}$ enjoys a similar spirit of recent clustering based unsupervised representation learning~\cite{cliquecnn2016,xie2016unsupervised,caron2018deep,yan2020clusterfit,caron2020unsupervised,li2020prototypical,asano2020self,van2020scan,tao2021idfd}.$_{\!}$  However,$_{\!}$ it$_{\!}$ works$_{\!}$ in$_{\!}$ a$_{\!}$ class-wise$_{\!}$ manner,$_{\!}$ since$_{\!}$ the$_{\!}$ class$_{\!}$ label$_{\!}$ is$_{\!}$ given.$_{\!}$ In$_{\!}$ this$_{\!}$ way,$_{\!}$ DNC$_{\!}$ opti- mizes the representation by adjusting the arrangement between sub-centroids and data samples. The enhanced representation in turn helps to find more informative$_{\!}$ sub-centroids,$_{\!}$ benefiting$_{\!}$ classification$_{\!}$ eventually.$_{\!}$ As$_{\!}$ such,$_{\!}$ DNC$_{\!}$~conducts \textit{unsupervised sub-class pattern discovery} during \textit{supervised representation learning}, distinguishing it from most (if not all)~of current visual recognition models.

Since the latent representation $f$ evolves continually during training, class~sub-centroids should be
synchronized, which requires performing class-wise clustering over~all training data after each batch

update. This is highly expensive on large datasets, even though Sinkhorn-Knopp iteration$_{\!}$~\cite{cuturi2013sinkhorn} based
clustering$_{\!}$ (\ref{eq:BIP2})$_{\!}$ is$_{\!}$ highly$_{\!}$ efficient. To circumvent the expensive, offline sub-centroid estimation, we adopt \textit{momentum$_{\!}$ update}$_{\!}$ and \textit{online} clustering.$_{\!}$ Specifically,$_{\!}$ at$_{\!}$ each$_{\!}$ training$_{\!}$ iteration,$_{\!}$~we$_{\!}$ conduct$_{\!}$ class-wise$_{\!}$ clustering$_{\!}$ on$_{\!}$ the$_{\!}$ current$_{\!}$ batch$_{\!}$ and$_{\!}$ update$_{\!}$ each$_{\!}$ sub-centroid$_{\!}$~as:$_{\!\!\!}$
\begin{equation}\label{eq:mom}
\small
\bm{p}^c_{k} \!\leftarrow\! \mu\bm{p}^c_{k} + (1-\mu)\bar{\bm{x}}^c_{k},
\end{equation}
where$_{\!}$ $\mu\!\in\![0, 1]_{\!}$ is$_{\!}$ a$_{\!}$ momentum$_{\!}$ coefficient,$_{\!}$ and$_{\!}$ $\bar{\bm{x}}^c_{k}\!\in\!\mathbb{R}^{d\!}$ is$_{\!}$ the$_{\!}$ mean$_{\!}$ feature$_{\!}$ vector$_{\!}$ of$_{\!}$ data$_{\!}$ assigned$_{\!}$~to $(c,k)$-cluster$_{\!}$ in$_{\!}$ current$_{\!}$ batch.$_{\!}$ As$_{\!}$ such,$_{\!}$ the$_{\!}$ sub-centroids$_{\!}$ can$_{\!}$ be$_{\!}$ up-to-date$_{\!}$ with$_{\!}$ the$_{\!}$ change$_{\!}$ of$_{\!}$ para- meters.$_{\!}$ Though$_{\!}$ effective$_{\!}$ enough$_{\!}$ in$_{\!}$ most$_{\!}$ cases,$_{\!}$ batch-wise$_{\!}$ clustering$_{\!}$ could$_{\!}$ not$_{\!}$ extend$_{\!}$ to$_{\!}$ a$_{\!}$ large$_{\!}$ num- ber of classes, \eg, when training on ImageNet$_{\!}$~\cite{ImageNet} with 1K classes using batch size 256, not~all the classes/clusters are present in a batch. To solve this, we store features~from several prior batches~in a
memory, and do clustering on both the memory and current batch. DNC can be trained by gradient backpropagation~in small-batch setting, with negligible lagging ($\sim$5\% training delay on ImageNet).



\noindent\textbf{Versatility.} DNC is a general framework; it can be effortless integrated into any parametric~classifier based DNNs, with minimal$_{\!}$ architecture change,$_{\!}$ \ie,$_{\!}$ removing$_{\!}$ the$_{\!}$ parametric softmax$_{\!}$ layer. However, $_{\!}$ DNC$_{\!}$ changes$_{\!}$ the$_{\!}$ classification$_{\!}$ decision-making$_{\!}$ mode,$_{\!}$  reforms$_{\!}$ the$_{\!}$ training regime, and$_{\!}$ makes$_{\!}$ the$_{\!}$ reasoning$_{\!}$ process$_{\!}$ more$_{\!}$ transparent,$_{\!}$ without$_{\!}$ slowing$_{\!}$ the$_{\!}$ inference$_{\!}$ speed.$_{\!}$ DNC$_{\!}$ can$_{\!}$~be ap- plied$_{\!}$ to$_{\!}$ various$_{\!}$ visual$_{\!}$ recognition$_{\!}$ tasks,$_{\!}$ including$_{\!}$ image$_{\!}$ classification$_{\!}$ (\S\ref{sec:cls})$_{\!}$ and$_{\!}$ segmentation$_{\!}$ (\S\ref{sec:seg}).

\noindent\textbf{Transferability.}$_{\!}$ As$_{\!}$ a$_{\!}$ nonparametric$_{\!}$ scheme,$_{\!}$ DNC can handle an$_{\!}$ arbitrary$_{\!}$ number$_{\!}$ $_{\!}$of$_{\!}$ classes$_{\!}$~with$_{\!}$ fixed$_{\!}$~output$_{\!}$ dimensionality$_{\!}$ ($d$);$_{\!}$ all$_{\!}$ the$_{\!}$ knowledge$_{\!}$ learnt$_{\!}$ on$_{\!}$ a$_{\!}$ source$_{\!}$ task$_{\!}$ (\eg,$_{\!}$ ImageNet$_{\!}$ classification$_{\!}$ with$_{\!}$ 1K$_{\!}$ classes)$_{\!}$~are$_{\!}$ stored$_{\!}$ as$_{\!}$ a$_{\!}$ constant$_{\!}$ amount$_{\!}$ of$_{\!}$ parameters$_{\!}$ in$_{\!\!}$ $f$,$_{\!}$ and$_{\!}$ thus$_{\!}$ can$_{\!}$ be$_{\!}$ \textit{completely}$_{\!}$ transferred for a new task (\eg, Cityscapes$_{\!}$~\cite{cordts2016cityscapes} segmentation with 19 classes), under the$_{\!}$ ``pre-training$_{\!}$ and$_{\!}$ fine-tuning''$_{\!}$ paradigm.$_{\!}$ In$_{\!}$ a$_{\!}$ similar$_{\!}$ setting,$_{\!}$ the parametric counterpart$_{\!}$ has$_{\!}$ to$_{\!}$ discard$_{\!}$ 2M$_{\!}$ parameters$_{\!}$ during$_{\!}$ transfer$_{\!}$ learning$_{\!}$ ($d_{\!}\!=_{\!}\!2048$ when$_{\!}$ using$_{\!}$ ResNet101$_{\!}$~\cite{he2016deep}). See \S\ref{sec:seg} for related experiments.

\noindent\textbf{Ad-hoc$_{\!}$ Explainability.$_{\!}$} DNC$_{\!}$ is$_{\!}$ a$_{\!}$ transparent$_{\!}$ classifier$_{\!}$ that$_{\!}$ has$_{\!}$ a$_{\!}$ built-in$_{\!}$ case-based reasoning process, as the sub-centroids are summarized from real observations and actually used during classification. So far we only discussed the case where the sub-centroids are considered as average mean feature vectors of a~few training samples with similar patterns. When restricting the sub-centroids to be elements of the training set  (\ie, representative training images), DNC naturally comes with human-understandable explanations for each prediction, and the explanations are loyal to the inter- nal$_{\!}$ decision$_{\!}$ mode$_{\!}$ and$_{\!}$ not$_{\!}$ created$_{\!}$~post-hoc.$_{\!}$ Studies$_{\!}$ regarding$_{\!}$ \textit{ad-hoc}$_{\!}$ explainability$_{\!}$ are$_{\!}$ given$_{\!}$ in$_{\!}$ \S\ref{sec:adex}.



\vspace{-10pt}
\section{Experiment}
\label{sec:exp}
\vspace{-6pt}

\subsection{Experiments on Image Classification}\label{sec:cls}
\vspace{-6pt}
\noindent\textbf{Dataset.} The evaluation for image classification is carried out on CIFAR-10\!~\cite{CIFAR} and ImageNet\!~\cite{ImageNet}.  


\noindent\textbf{$_{\!}$Network$_{\!}$ Architecture.}$_{\!}$ For$_{\!}$ completeness,$_{\!}$ we$_{\!}$ craft$_{\!}$ DNC$_{\!}$ on$_{\!}$~popular CNN-based ResNet50/100$_{\!}$~\cite{he2016deep} and recent Transformer-based Swin-Small/-Base$_{\!}$~\cite{liu2021swin}. Note that, we only remove the last linear classification layer, and the final output dimensionality of DNC is as many as the last layer feature of the parametric counterpart, \ie, 2,048 for ResNet50/100, 768 for Swin-Small, and 1,024 for Swin-Base.

\noindent\textbf{Training.$_{\!}$} We$_{\!}$ use$_{\!}$  \texttt{mmclassification}\footnote{https://github.com/open-mmlab/mmclassification}$_{\!}$ as$_{\!}$ codebase$_{\!}$ and$_{\!}$ follow$_{\!}$ the$_{\!}$ \textit{default}$_{\!}$~training$_{\!}$ settings.$_{\!}$ For$_{\!}$ CIFAR-10, we train ResNet for $200$ epochs, with batch size~$128$. The$_{\!}$ memory$_{\!}$ size$_{\!}$ for$_{\!}$ DNC$_{\!}$ models$_{\!}$ is$_{\!}$ set$_{\!}$ as$_{\!}$ 100$_{\!}$ batches.$_{\!}$ For$_{\!}$ ImageNet,$_{\!}$ we$_{\!}$ train$_{\!}$ 100$_{\!}$ and$_{\!}$ 300$_{\!}$ epochs with batch size $16$ for ResNet and Swin, respectively. The initial learning rates of ResNet and Swin are set as $0.1$ and $0.0005$, scheduled by a step policy and polynomial annealing  policy, respectively. Limited by our~GPU capacity,$_{\!}$ the$_{\!}$ memory$_{\!}$ sizes$_{\!}$ are$_{\!}$ set$_{\!}$ as$_{\!}$ $1,\!000$$_{\!}$ and$_{\!}$ $500$$_{\!}$ batches$_{\!}$ for$_{\!}$ DNC$_{\!}$ versions$_{\!}$ of$_{\!}$ ResNet$_{\!}$ and$_{\!}$ Swin,$_{\!}$ respectively.$_{\!}$ Other$_{\!}$  hyper-
\setlength\intextsep{5pt}
\begin{wraptable}{r}{0.45\linewidth}
\vspace{-1pt}
	\setlength{\abovecaptionskip}{0cm}
	\caption{\textbf{$_{\!\!\!}$Classification$_{\!}$ \texttt{top-1}$_{\!}$ accuracy$_{\!}$} on$_{\!}$ CIFAR-10$_{\!}$~\cite{CIFAR}$_{\!}$ \texttt{test}.$_{\!}$ {\#Params:$_{\!}$ the$_{\!}$ number$_{\!}$~of$_{\!}$ learnable$_{\!}$ parameters$_{\!}$ (same$_{\!}$ for$_{\!}$ other$_{\!}$ tables).$_{\!}$}}
\hspace{-0.8em}
	\resizebox{0.45\textwidth}{!}{
		\setlength\tabcolsep{4pt}
		\renewcommand\arraystretch{1}
		\begin{tabular}{|rc||cc| }
			\hline\thickhline
			\rowcolor{mygray}
			Method~~~~ & Backbone & \#Params& \texttt{top-1} \\	\hline	\hline 
			ResNet \cite{he2016deep} &\multirow{2}{*}{ResNet50} & 23.52M & 95.55\% \\
            \textbf{DNC}-ResNet & & 23.50M &\textbf{95.78}\%\\\hline
			ResNet \cite{he2016deep} &\multirow{2}{*}{ResNet101} & 42.51M & 95.58\%\\  
            \textbf{DNC}-ResNet & &42.49M & \textbf{95.82}\%\\
			\hline
	\end{tabular}}
	\label{table:cifar10}
\vspace{-10pt}
\end{wraptable}
parameters$_{\!}$ are$_{\!}$ empirically$_{\!}$ set$_{\!}$ as:$_{\!}$ $K_{\!}\!=\!4$$_{\!}$ and$_{\!}$ $\mu\!=\!0.999$.$_{\!\!\!}$ Models$_{\!}$ are$_{\!}$ trained$_{\!}$ \textit{from$_{\!}$ scratch}$_{\!}$ on$_{\!}$ eight$_{\!}$ V100$_{\!}$~GPUs.$_{\!}$



\noindent\textbf{Results$_{\!}$ on$_{\!}$ CIFAR-10.}$_{\!}$ Table$_{\!}$~\ref{table:cifar10}$_{\!}$ compares$_{\!}$ DNC$_{\!}$~with$_{\!}$ the parametric$_{\!}$ counterpart,$_{\!}$ based$_{\!}$ on$_{\!}$ the$_{\!}$ most$_{\!}$ repre- sentative$_{\!}$ CNN$_{\!}$ network$_{\!}$ architecture,$_{\!}$ \ie,$_{\!}$ ResNet.$_{\!}$ As$_{\!}$ seen,$_{\!}$~DNC$_{\!}$ obtains$_{\!}$ better$_{\!}$ performance:$_{\!}$ DNC$_{\!}$ is$_{\!}$ \textbf{0.23}\%$_{\!}$
higher$_{\!}$~on$_{\!}$ ResNet50,$_{\!}$ and$_{\!}$ \textbf{0.24}\%$_{\!}$ higher$_{\!}$ on$_{\!}$ ResNet101,$_{\!}$  using$_{\!}$ fewer$_{\!}$ learnable$_{\!}$ parameters.$_{\!}$ With$_{\!}$ the$_{\!}$ exact$_{\!}$ same$_{\!}$ backbone$_{\!}$ architectures$_{\!}$ and$_{\!}$ training$_{\!}$ settings,$_{\!}$ one$_{\!}$ can$_{\!}$ safely$_{\!}$ attribute$_{\!}$ the$_{\!}$ performance$_{\!}$ gain$_{\!}$ to$_{\!}$ DNC.


\setlength\intextsep{5pt}
\begin{wraptable}{r}{0.5\linewidth}
	\centering
\vspace{5pt}
	\setlength{\abovecaptionskip}{0cm}
	\setlength{\belowcaptionskip}{-0.2cm}
	\caption{\textbf{Classification$_{\!}$ \texttt{top-1}$_{\!}$ and \texttt{top-5}$_{\!}$ accuracy} on ImageNet~\cite{ImageNet} \texttt{val}.}
\hspace{-0.6em}
	\resizebox{0.5\textwidth}{!}{
		\setlength\tabcolsep{2pt}
		\renewcommand\arraystretch{1}
		\begin{tabular}{|rc||ccc|}
			\hline\thickhline
			\rowcolor{mygray}
			Method~~~~ & Backbone & \#Params & \texttt{top-1} & \texttt{top-5}  \\	\hline	\hline
			ResNet \cite{he2016deep} &\multirow{2}{*}{ResNet50} & 25.56M & 76.20\%& 93.01\%\\
            \textbf{DNC}-ResNet& &23.51M& \textbf{76.49}\% & \textbf{93.08}\%\\\hline
			ResNet \cite{he2016deep} &\multirow{2}{*}{ResNet101} & 44.55M & 77.52\%& 93.06\%\\
			\textbf{DNC}-ResNet& &42.50M& \textbf{77.80}\% & \textbf{93.85}\% \\\hline
            Swin \cite{liu2021swin} &\multirow{2}{*}{Swin-S} & 49.61M & 83.02\%& 96.29\%\\
            \textbf{DNC}-Swin& &48.84M& \textbf{83.26}\%& \textbf{96.40}\% \\\hline
            Swin \cite{liu2021swin} &\multirow{2}{*}{Swin-B} & 87.77M & 83.36\%& 96.44\%\\
            \textbf{DNC}-Swin& &86.75M& \textbf{83.68}\%& \textbf{97.02}\%\\ \hline
	\end{tabular}}
	\label{table:imagenet}
\end{wraptable}
\noindent\textbf{Results$_{\!}$ on$_{\!}$ ImageNet.}$_{\!}$ Table$_{\!}$~\ref{table:imagenet}$_{\!}$ illustrates$_{\!}$ again$_{\!}$ our$_{\!}$ compelling$_{\!}$ results$_{\!}$ over$_{\!}$ different$_{\!}$ vision$_{\!}$~net-$_{\!}$  work$_{\!}$ architectures.$_{\!}$ In$_{\!}$ terms$_{\!}$ of \texttt{top-1}$_{\!}$ acc.,$_{\!}$ our

\vspace{-6.5pt}
DNC$_{\!}$ exceeds$_{\!}$ the$_{\!}$ parametric$_{\!}$  classifier$_{\!}$ by$_{\!}$ \textbf{0.29}\%$_{\!}$ and$_{\!}$ \textbf{0.28}\%$_{\!}$
on$_{\!}$~ResNet50$_{\!}$ and$_{\!}$ ResNet101,$_{\!}$ respec- tively. DNC also obtains promising$_{\!}$ results$_{\!}$ with$_{\!}$ Transformer  architecture:$_{\!}$ \textbf{83.26}\%$_{\!}$ \textit{vs}$_{\!}$ 83.02\%$_{\!}$ ~on Swin-S, \textbf{83.68}\% \textit{vs} 83.36\% on Swin-B. These results are impressive,$_{\!}$ considering$_{\!}$ the transpa- rent, case-based reasoning nature of DNC.~One may$_{\!}$ notice$_{\!}$ another$_{\!}$ nonparametric,$_{\!}$ $k$-NN$_{\!}$~classi- fier \cite{wu2018improving} reports 76.57\%$_{\!}$ \texttt{top-1}$_{\!}$~acc.,$_{\!}$ based$_{\!}$ on$_{\!}$~ResNet50.$_{\!}$ However,$_{\!}$ \cite{wu2018improving}$_{\!}$ is$_{\!}$ trained$_{\!}$ with 130 epochs;~in\\
\noindent this$_{\!}$ setting,$_{\!}$ DNC$_{\!}$ gains$_{\!}$ 76.64\%.$_{\!}$ Moreover,$_{\!}$ as$_{\!}$ mentioned$_{\!}$ in$_{\!}$ \S\ref{sec:rw},$_{\!}$ \cite{wu2018improving}$_{\!}$ poses$_{\!}$ huge$_{\!}$ storage$_{\!}$ demand,$_{\!}$ \ie,$_{\!}$~retaining$_{\!}$ the$_{\!}$ whole$_{\!}$ ImageNet$_{\!}$ training$_{\!}$ set$_{\!}$ (\ie,$_{\!}$ 1.2M$_{\!}$ images)$_{\!}$ to$_{\!}$ perform$_{\!}$ the$_{\!}$ $k$-NN$_{\!}$~decision rule, and suffers from very low efficiency, caused by extensive comparisons between each test image and all the training images. These limitations prevent$_{\!}$ the$_{\!}$ adoption$_{\!}$ of$_{\!}$ \cite{wu2018improving}$_{\!}$ in$_{\!}$ real$_{\!}$ application$_{\!}$ scenarios.$_{\!}$~In$_{\!}$ contrast,$_{\!}$ DNC$_{\!}$ only$_{\!}$ relies$_{\!}$ on$_{\!}$ a small set of class representatives (\ie,$_{\!}$~four sub-centroids per class) for\\
\noindent  classification decision-making and causes no extra computation budget during network deployment. 



\vspace{-5pt}
\subsection{Experiments on Semantic Segmentation} \label{sec:seg}
\vspace{-3pt}
\noindent\textbf{Dataset.}$_{\!}$ The$_{\!}$ evaluation$_{\!}$ for$_{\!}$ semantic$_{\!}$ segmentation$_{\!}$ is$_{\!}$ carried$_{\!}$ out$_{\!}$ on$_{\!}$ ADE20K$_{\!}$~\cite{zhou2017scene}$_{\!}$ and$_{\!}$ Cityscapes$_{\!}$~\cite{cordts2016cityscapes}.$_{\!}$  



\noindent\textbf{Segmentation Network Architecture.} For comprehensive evaluation, we approach DNC on three famous$_{\!}$ segmentation$_{\!}$ models$_{\!}$ (\ie,$_{\!}$ FCN$_{\!}$~\cite{long2015fully},$_{\!}$ DeepLab$_\text{V3}$$_{\!\!}$~\cite{chen2017rethinking},$_{\!}$ UperNet$_{\!}$~\cite{xiao2018unified}),$_{\!}$ using$_{\!}$ two$_{\!}$ backbone architectures (\ie, ResNet101$_{\!}$~\cite{he2016deep} and Swin-B$_{\!}$~\cite{liu2021swin}).$_{\!}$ For$_{\!}$ the$_{\!}$ segmentation$_{\!}$ models,$_{\!}$ the$_{\!}$ only$_{\!}$ architec- tural$_{\!}$ modification is the removal$_{\!}$ of$_{\!}$ the$_{\!}$ ``segmentation$_{\!}$ head''$_{\!}$ (\ie,$_{\!}$ $1\!\times\!1$$_{\!}$ conv$_{\!}$ based,$_{\!}$  pixel-wise$_{\!}$ classi- fication$_{\!}$ layer).$_{\!}$ For$_{\!}$ the$_{\!}$ backbone$_{\!}$ networks,$_{\!}$~we$_{\!}$ respectively$_{\!}$ adopt$_{\!}$ parametric$_{\!}$ classifier based and our nonparametric, DNC based versions, which are both trained~on ImageNet$_{\!}$~\cite{ImageNet} and reported in Table$_{\!}$~\ref{table:imagenet}, for initialization. Thus for each segmentation model, we derive four variants from the different combinations of parametric and DNC versions of the backbone and segmentation network architectures.

\noindent\textbf{Training.} We adopt \texttt{mmsegmentation}\footnote{https://github.com/open-mmlab/mmsegmentation} as the codebase, and follow the \textit{default} training settings.\\
\noindent We train FCN and DeepLab$_\text{V3}$ with ResNet101 using SGD~optimizer with an initial learning rate~$0.1$, and UperNet with Swin-B using~AdamW with an initial learning rate $\text{6e-5}$. For all the models, the learning rate is scheduled following a polynomial annealing policy. As common practices \cite{xie2021segformer,cheng2021maskformer}, we$_{\!}$ train$_{\!}$ the$_{\!}$ models$_{\!}$ on$_{\!}$ ADE20K$_{\!}$ \texttt{train}$_{\!}$  with$_{\!}$  crop$_{\!}$ size$_{\!}$ $512\!\times\!512$$_{\!}$ and$_{\!}$ batch$_{\!}$ size$_{\!}$ $16$; on Cityscapes \texttt{train} with crop size $769\!\times\!769$ and batch size $8$.  All the models~are trained for $160$K iterations on\\
\noindent both datasets.$_{\!}$ 
Standard$_{\!}$ data$_{\!}$ augmentation$_{\!}$ techniques$_{\!}$ are$_{\!}$ used,$_{\!}$ including$_{\!}$~scale and color jittering, flipping, and cropping.$_{\!}$ The$_{\!}$ hyper-parameters$_{\!}$ of$_{\!}$ DNC$_{\!}$ are$_{\!}$ by$_{\!}$ default$_{\!}$ set as:$_{\!}$ $K\!=\!10$$_{\!}$ and$_{\!}$ $\mu\!=\!0.999$.

\setlength\intextsep{5pt}
\begin{wraptable}{r}{0.6\linewidth}
	\centering
\vspace{0pt}
	\setlength{\abovecaptionskip}{0cm}
	\setlength{\belowcaptionskip}{-0.2cm}
	\caption{\textbf{Segmentation mIoU score} on ADE20K~\cite{zhou2017scene} \texttt{val} and Cityscapes \cite{cordts2016cityscapes} \texttt{val} ($\texttt{top-1}$ acc. on ImageNet$_{\!}$~\cite{ImageNet} \texttt{val} of backbones are also reported for reference).}
\hspace{-0.6em}
	\resizebox{0.6\columnwidth}{!}{
		\setlength\tabcolsep{1pt}
		\renewcommand\arraystretch{1.05}
		\begin{tabular}{|r|cc||ccc|}
			\hline\thickhline
			\rowcolor{mygray}
			&   &${\!\!\!\!\!\!}$ImageNet  & &{ADE20K} & {Cityscapes} \\
			\rowcolor{mygray}
			\multirow{-2}{*}{Method~~~~~~}& \multirow{-2}{*}{Backbone} &{\texttt{top-1} acc.}  & \multirow{-2}{*}{\#Params} & \texttt{mIoU}& \texttt{mIoU} \\ \hline\hline
				
			&ResNet101~\cite{he2016deep}
			& 77.52\% & 68.6M &39.9\% & 75.6\% \\
						
			\multirow{-2}{*}{FCN~\cite{long2015fully}}
			&\textbf{DNC}-ResNet101 &77.80\% &68.6M & \reshl{40.4\%}{0.5} & \reshl{76.3\%}{0.7} \\

			&ResNet101~\cite{he2016deep}
			& 77.52\%   &68.5M & \reshl{41.1\%}{1.2} & \reshl{76.7\%}{1.1} \\
			
			\multirow{-2}{*}{\textbf{DNC}-FCN} &\textbf{DNC}-ResNet101
			& 77.80\% &68.5M & \reshx{\textbf{42.3}\%}{2.4}   & \reshx{\textbf{77.5}\%}{1.9} \\ \hline

			&ResNet101~\cite{he2016deep}
			& 77.52\% & 62.7M &44.1\% & 78.1\%\\
			
			\multirow{-2}{*}{DeepLab$_\text{V3}$~\cite{chen2017rethinking}}
			&\textbf{DNC}-ResNet101  &77.80\% &62.7M & \reshl{44.6\%}{0.5} & \reshl{78.7\%}{0.6} \\

			&ResNet101~\cite{he2016deep}
			& 77.52\% &62.6M & \reshl{45.0\%}{0.9}  & \reshl{79.1\%}{1.0} \\
			
			\multirow{-2}{*}{\textbf{DNC}-DeepLab$_\text{V3}$}
			&\textbf{DNC}-ResNet101 & 77.80\% &62.6M & \reshx{\textbf{45.7}\%}{1.6} & \reshx{\textbf{79.8}\%}{1.7} \\ \hline

			&Swin-B~\cite{liu2021swin}
			& 83.36\% &90.6M & 48.0\% & 79.8\% \\

			\multirow{-2}{*}{UperNet~\cite{xiao2018unified}}
			&\textbf{DNC}-Swin-B &83.68\%  &90.6M  & \reshl{48.4\%}{0.4}  & \reshl{80.1\%}{0.3} \\

			&Swin-B~\cite{liu2021swin}
			&83.36\% &90.5M & \reshl{48.6\%}{0.6} & \reshl{80.5\%}{0.7} \\
			
			\multirow{-2}{*}{\textbf{DNC}-UperNet}&\textbf{DNC}-Swin-B
			&83.68\% &90.5M & \reshx{\textbf{50.5}\%}{2.5} & \reshx{\textbf{80.9}\%}{1.1} \\ \hline
	\end{tabular}}
	\label{table:seg}
\end{wraptable}

\noindent\textbf{Performance on Segmentation.}~As summarized$_{\!}$ in$_{\!}$ Table$_{\!}$ \ref{table:seg},$_{\!}$ our$_{\!}$ DNC$_{\!}$~seg-$_{\!}$ mentation$_{\!}$ models,$_{\!}$ no$_{\!}$ matter$_{\!}$ whether using$_{\!}$ DNC$_{\!}$ based$_{\!}$ backbones,$_{\!}$ obtain better$_{\!}$ performance$_{\!}$ over$_{\!}$ parametric competitors, \ie, FCN + ResNet101, DeepLab$_\text{V3}$ + ResNet101,$_{\!}$ UperNet + Swin-B,$_{\!}$ across$_{\!}$ different segmentation$_{\!}$ and$_{\!}$ backbone$_{\!}$ network$_{\!}$ architectures.$_{\!}$ Taking$_{\!}$ as$_{\!}$ an$_{\!}$ example$_{\!}$ the$_{\!}$ famous$_{\!}$ FCN,$_{\!\!\!}$ the$_{\!}$ original$_{\!}$ version$_{\!}$ gains$_{\!}$ 39.91\%$_{\!}$~and 75.6\% \texttt{mIoU} on ADE20K and City-  scapes,$_{\!}$ respectively.$_{\!}$ By$_{\!}$ comparison,$_{\!}$ with$_{\!}$ the$_{\!}$ same$_{\!}$ backbone$_{\!}$ --$_{\!}$ ResNet101,$_{\!}$
DNC-FCN$_{\!}$ boosts$_{\!}$ the$_{\!}$ scores$_{\!}$ to$_{\!}$ 41.1\%$_{\!}$ and$_{\!}$~76.7\%. When turning~to DNC pre-trained backbone -- {DNC}-ResNet101, our DNC-FCN outperforms again its parametric counterpart -- FCN, \ie, 42.3\% \textit{vs} 40.4\% on$_{\!}$ {ADE20K},$_{\!}$ 77.5\%$_{\!}$ \textit{vs}$_{\!}$ 76.3\%$_{\!}$  on$_{\!}$ {Cityscapes}.$_{\!}$ Similar$_{\!}$ trends$_{\!}$ can$_{\!}$ be$_{\!}$ also observed$_{\!}$~for DeepLab$_\text{V3}$ and UperNet. 

\noindent\textbf{Analysis on Transferability.} One appealing feature of DNC is its strong transferability, as DNC learns classification by directly comparing data samples in the feature space. 
The results in Table~\ref{table:seg}\\
\noindent also evidence this point. For example,
DeepLab$_\text{V3}$ -- even a parametric classifier based segmentation$_{\!}$ model$_{\!}$ --$_{\!}$ can$_{\!}$
 achieve$_{\!}$ large$_{\!}$ performance$_{\!}$ gains,$_{\!}$ \ie,$_{\!}$ 44.6\%$_{\!}$ \textit{vs}$_{\!}$ 44.1\%$_{\!}$ on$_{\!}$ ADE20K,$_{\!}$
 78.7\%$_{\!}$ \textit{vs}$_{\!}$ 78.1\%$_{\!}$ on Cityscapes, after using {DNC}-ResNet101 for fine-tuning. When it comes to DNC-DeepLab$_\text{V3}$, better$_{\!}$ performance$_{\!}$ can$_{\!}$ be$_{\!}$ achieved,$_{\!}$ after$_{\!}$ replacing$_{\!}$ ResNet101$_{\!}$ with$_{\!}$ {DNC}-ResNet101,$_{\!}$
 \ie,$_{\!}$ 45.7\%$_{\!}$ \textit{vs}$_{\!}$ 45.0\% on ADE20K, 79.8\% \textit{vs} 79.1\% on Cityscapes. We speculate this is because, when the segmentation$_{\!}$ and backbone networks are both~built upon DNC, the model only needs to adapt the original representation space to the target task, without learning any extra new parameters. Also, these impressive results reflect the innovative opportunity of applying our DNC for more downstream visual recognition tasks, either as a new classification network architecture or a strong and transferable backbone.

\vspace{-10pt}
\subsection{Study of Ad-hoc Explainability}\label{sec:adex}
\vspace{-6pt}
So far, we have empirically showed that, by inheriting the intuitive power of Nearest Centroids~and the$_{\!}$ strong$_{\!}$ representation$_{\!}$ learning$_{\!}$ ability$_{\!}$ of$_{\!}$ DNNs,$_{\!}$ DNC$_{\!}$ can$_{\!}$
 serve$_{\!}$ as$_{\!}$ a$_{\!}$ transparent$_{\!}$ yet$_{\!}$ powerful$_{\!}$ tool for~visual recognition tasks~with improved transferability.
 When anchoring the class sub-centroids~to\\
\noindent real observations (\ie, actual images from the training dataset), instead of selecting them as cluster centers (\ie, mean features of a set of training images), one may expect that DNC will gain enhanced \textit{ad-hoc} explainability. Next we will show this is indeed possible, only at negligible performance cost.

\vspace{-2pt}
\noindent\textbf{Experimental$_{\!}$ Setup.}$_{\!}$  We$_{\!}$  still$_{\!}$  train$_{\!}$  DNC$_{\!}$
 version$_{\!}$  of$_{\!}$  ResNet50$_{\!}$  on$_{\!}$  ImageNet$_{\!}$ \texttt{train}$_{\!}$ for$_{\!}$ 100$_{\!}$ epochs.$_{\!}$ In$_{\!}$ the$_{\!}$ first$_{\!}$ 90$_{\!}$ epochs,$_{\!}$ the$_{\!}$
model$_{\!}$ is$_{\!}$ trained$_{\!}$ in$_{\!}$~a$_{\!}$ standard$_{\!}$ manner,$_{\!}$ \ie,$_{\!}$
computing$_{\!}$ the$_{\!}$ class$_{\!}$
sub-centroids$_{\!}$ as$_{\!}$ cluster$_{\!}$ centers.$_{\!}$ Then$_{\!}$
we anchor each sub-centroid to its closest training image, based  on the~co- sine similarity of features. In
the  final 10  epochs, the sub-centroids are only updated as the features
\setlength\intextsep{5pt}
\begin{wraptable}{r}{0.715\linewidth}
	\centering
\vspace{-2pt}
	\setlength{\abovecaptionskip}{0cm}
	\setlength{\belowcaptionskip}{-0.2cm}
  \centering
      \includegraphics[width=0.99 \linewidth]{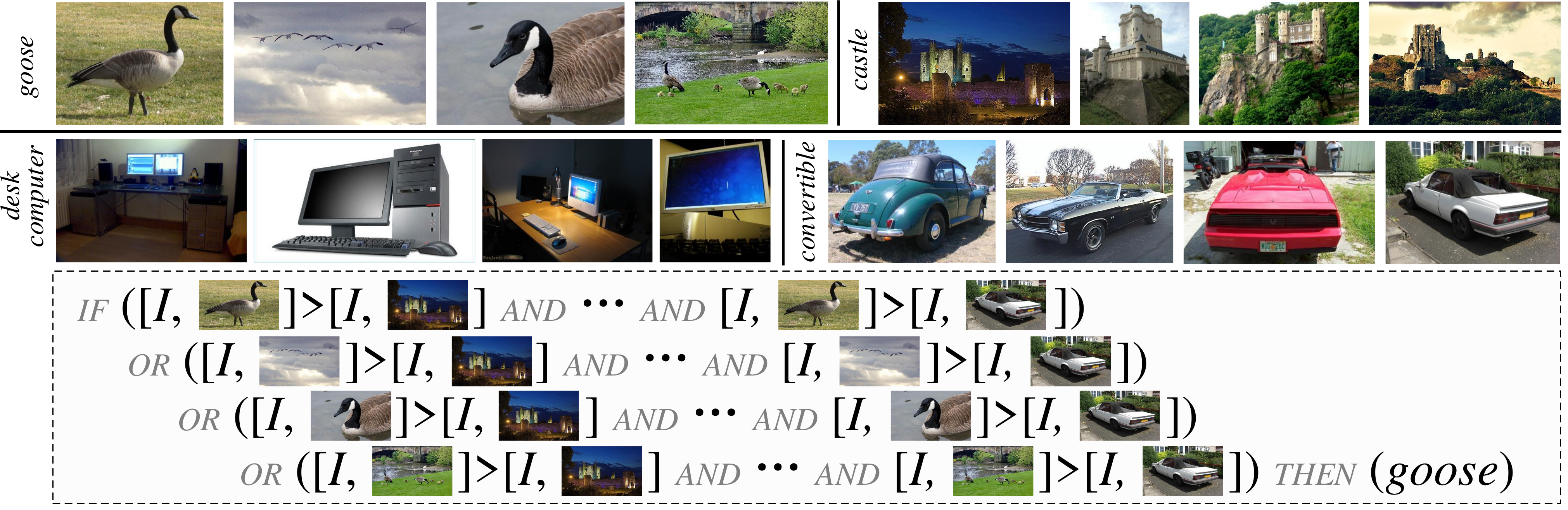}
     \vspace{2pt}
     \makeatletter\def\@captype{figure}\makeatother
\caption{$_{\!\!}$\textit{Top:}$_{\!}$ sub-centroid$_{\!}$ images.$_{\!}$ \textit{Bottom:}$_{\!}$ rule$_{\!}$ created$_{\!}$ for$_{\!}$ ``goose''.}
\label{fig:vs}
\end{wraptable}
of~their~anchored$_{\!}$ training$_{\!}$ images.$_{\!}$~Besides$_{\!}$~this,~all the$_{\!}$~other$_{\!}$~training$_{\!}$~settings$_{\!}$
are$_{\!}$~as$_{\!}$~normal.$_{\!}$~In$_{\!}$~this$_{\!}$ way, we can get a more interpretable DNC-ResNet50.

\vspace{-2pt}
\noindent\textbf{$_{\!}$Interpretable$_{\!}$~Class$_{\!}$~Sub-$_{\!\!\!}$}
\textbf{$_{\!}$centroids.}$_{\!}$~The$_{\!}$~top$_{\!}$~of$_{\!}$~Fig.$_{\!\!}$~\ref{fig:vs}$_{\!\!}$
$_{\!}$plots$_{\!}$ our$_{\!}$ discovered$_{\!}$ sub-centroid images for four
ImageNet classes. These representative images are diverse in appearance, viewpoints, illuminations, scales, \textit{etc.},~characterizing their$_{\!}$ corresponding$_{\!}$ classes$_{\!}$ and$_{\!}$ allowing$_{\!}$ humans$_{\!}$ to$_{\!}$ view$_{\!}$ and$_{\!}$ understand.

\setlength\intextsep{0pt}
\begin{wraptable}{r}{0.47\linewidth}
	\centering
\vspace{-0.2em}
	\setlength{\abovecaptionskip}{0cm}
	\caption{\textbf{Classification$_{\!}$ \texttt{top-1}$_{\!}$ and~\texttt{top-5}$_{\!}$~ac- curacy}$_{\!\!}$ on$_{\!\!}$ ImageNet$_{\!\!}$~\cite{ImageNet}$_{\!\!}$ \texttt{val},$_{\!\!}$ using$_{\!\!}$ cluster$_{\!\!}$~cen-$_{\!\!\!}$ ter \textit{vs} resembling real observation as class sub-centroids,$_{\!}$ based$_{\!}$ on$_{\!}$ {DNC}-ResNet50$_{\!}$ architecture.$_{\!\!\!}$}
\hspace{-0.9em}
	\resizebox{0.47\textwidth}{!}{
		\setlength\tabcolsep{3pt}
		\renewcommand\arraystretch{1}
		\begin{tabular}{|rc||cc|}
			\hline\thickhline
			\rowcolor{mygray}
			Sub-centroid~~ & Architecture & \texttt{top-1} & \texttt{top-5}  \\	\hline	\hline
            \textit{cluster center} &\multirow{2}{*}{\textbf{DNC}-ResNet50} &{76.49}\% &{93.08}\%\\
            \textit{real observation}  & &{76.37}\% &{93.04}\%\\\hline
            -~~~~~~~~ & ResNet50\cite{he2016deep}& 76.20\%& 93.01\%\\\hline
	\end{tabular}}
	\label{table:eximagenet}
\end{wraptable}

\vspace{-2pt}
\noindent\textbf{$_{\!}$Performance$_{\!}$ with$_{\!}$ Improved$_{\!}$ Interpretability.$_{\!\!}$}~We$_{\!}$ then$_{\!}$ report$_{\!}$ the$_{\!}$ score$_{\!}$ of$_{\!}$ our$_{\!}$ DNC-ResNet50$_{\!}$ based$_{\!}$ on$_{\!}$ the$_{\!}$ interpretable$_{\!}$ class$_{\!}$ representatives$_{\!}$ on$_{\!}$ Image- Net \texttt{val}. As shown in Table$_{\!}$~\ref{table:eximagenet},$_{\!}$~enforcing$_{\!}$ the$_{\!}$ class$_{\!}$
sub-centroids$_{\!}$ as$_{\!}$ real$_{\!}$ training$_{\!}$ images$_{\!}$ only brings marginal$_{\!}$ performance$_{\!}$ degradation$_{\!}$ (\eg,$_{\!}$ 76.49\%$\rightarrow$
76.37\%$_{\!}$ \texttt{top-1}$_{\!}$ acc.),$_{\!}$ while$_{\!}$ coming$_{\!}$ with$_{\!}$~better$_{\!}$ in- terpretability.$_{\!}$ More$_{\!}$ impressively,$_{\!}$ \textit{\textbf{our$_{\!}$ explainable$_{\!\!\!}$ DNC-ResNet50$_{\!}$ even$_{\!}$ outperforms the$_{\!}$ vanilla$_{\!}$ black-box$_{\!}$ ResNet50}},$_{\!}$ \eg,$_{\!}$ 76.37\%$_{\!}$ \textit{vs}$_{\!}$ 76.20\%.

\vspace{-2pt}
\noindent\textbf{Explain$_{\!}$ Inner$_{\!}$ Decision-Making$_{\!}$ Mode$_{\!}$ based$_{\!}$ on$_{\!}$ \textit{IF}$_{\!}$ $\cdots$ $_{\!\!\!}$\textit{Then}$_{\!}$ Rules.$_{\!}$} With$_{\!}$ the$_{\!}$ simple$_{\!}$ Nearest$_{\!}$ Cen- troids$_{\!}$ mechanism,$_{\!}$ we$_{\!}$~can use$_{\!}$ the$_{\!}$ representative$_{\!}$ images$_{\!}$ to$_{\!}$ form$_{\!}$ a$_{\!}$ set$_{\!}$ of \textit{IF}$_{\!}$ $\cdots$ $_{\!\!\!}$\textit{Then} rules$_{\!}$~\cite{angelov2020towards},$_{\!}$ so$_{\!}$ as$_{\!}$ to$_{\!}$ intuitively interpret the inner decision-making mode~of DNC for human users. In particular, let $\hat{I}$ denote a sub-centroid image for class $c$, $\check{I}_{1:T}$ representative images for all the other classes, and $I$ a query image. One linguistic$_{\!}$ logical \textit{IF}$_{\!}$ $\cdots$ $_{\!\!\!}$\textit{Then} rule can be generated for $\hat{I}$:
\begin{equation}
\small
\!\!\textit{IF}~\big([I, \hat{I}]_{\!}>_{\!}[I, \check{I}_{1}]~\textit{AND}~[I, \check{I}]_{\!}>_{\!}[I, \check{I}_{2}]~\textit{AND}\cdots\textit{AND}~[I, \hat{I}]_{\!}>_{\!}[I, \check{I}_{T}]\big)~\textit{THEN}~(\text{class}~c),
\end{equation}
where $[\cdot,\cdot]$ stands for similarity, given by DNC. The final rule for class $c$ is~created by$_{\!}$ combining$_{\!}$ all$_{\!}$ the$_{\!}$ rules$_{\!}$ of$_{\!}$ $K$$_{\!}$ sub-centroid$_{\!}$ images$_{\!}$ $\hat{I}_{1:K\!}$ of$_{\!}$ class$_{\!}$ $c$$_{\!}$ (see Fig.$_{\!}$~\ref{fig:vs}$_{\!}$ bottom):$_{\!}$
\begin{equation}
\small
\begin{aligned}\small
\textit{IF}~&\big([I, \hat{I}_{1}]_{\!}>_{\!}[I, \check{I}_{1}]~\textit{AND}\cdots\textit{AND}~[I, \hat{I}_{1}]_{\!}>_{\!}[I, \check{I}_{T}]\big)\\
&\textit{OR}~\big([I, \hat{I}_{2}]_{\!}>_{\!}[I, \check{I}_{1}]~\textit{AND}\cdots\textit{AND}~[I, \hat{I}_{2}]_{\!}>_{\!}[I, \check{I}_{T}]\big)\\
&\textit{OR}~\cdots\textit{OR}~\big([I, \hat{I}_{K}]_{\!}>_{\!}[I, \check{I}_{1}]~\textit{AND}\cdots\textit{AND}~[I, \hat{I}_{K}]_{\!}>_{\!}[I, \check{I}_{T}]\big)
~\textit{THEN}~(\text{class}~c).
\end{aligned}
\vspace{-12pt}
\end{equation}

\vspace{-10pt}
\begin{figure*}[t]
  \centering
       \vspace{-16pt}
      \includegraphics[width=1 \linewidth]{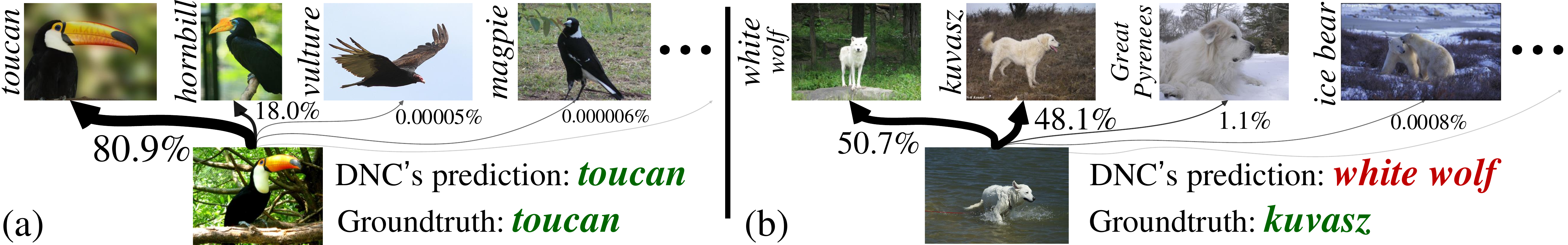}
     \vspace{-17pt}
\caption{$_{\!}$DNC$_{\!}$ can$_{\!}$ provide$_{\!}$ (dis)similarity-based$_{\!}$ interpretation.$_{\!}$ For$_{\!}$ the$_{\!}$ two$_{\!}$ test$_{\!}$ samples,$_{\!}$ we$_{\!}$ only$_{\!}$ plot the normalized similarities for their corresponding closest sub-centroids from top-4 scoring classes. }
\label{fig:ve}
\vspace{-15pt}
\end{figure*}

\vspace{-8pt}
\noindent\textbf{Interpret$_{\!}$ Prediction Based$_{\!}$ on$_{\!}$ (Dis)similarity to Sub-centroid Images.} Based$_{\!}$ on$_{\!}$ the$_{\!}$ interpretable$_{\!}$ class$_{\!}$ representatives,$_{\!}$ DNC$_{\!}$ can$_{\!}$ explain$_{\!}$ its$_{\!}$ predictions by letting users view and verify its computed (dis)similarity$_{\!}$ between$_{\!}$ query$_{\!}$ and$_{\!}$ class$_{\!}$ sub-centroid$_{\!}$ images.$_{\!}$
As$_{\!}$ shown$_{\!}$ in$_{\!}$ Fig.$_{\!}$~\ref{fig:ve}(a),$_{\!}$ an$_{\!}$ observation$_{\!}$ is correctly classified, as DNC thinks it looks (more) like a particular  exemplar of ``\textit{toucan}''. However, in$_{\!}$ Fig.$_{\!}$~\ref{fig:ve}(b),$_{\!}$ DNC$_{\!}$ struggles$_{\!}$ to$_{\!}$
 assign$_{\!}$ the$_{\!}$ observation$_{\!}$ to$_{\!}$ two$_{\!}$ exemplars$_{\!}$ from$_{\!}$ ``\textit{white$_{\!}$ wolf}''$_{\!}$ and ``\textit{kuvasz}'' respectively, and makes a wrong decision$_{\!}$ finally.$_{\!}$ Though$_{\!}$ users$_{\!}$ are$_{\!}$ unclear$_{\!}$ how$_{\!}$ DNC$_{\!}$ maps$_{\!}$ an$_{\!}$ image$_{\!}$~to$_{\!}$ feature, they can easily understand the decision-making mode$_{\!}$~\cite{rudin2022interpretable} (\eg, why is one class~predicted over another), and verify the calculated (dis)similarity -- the evidence for classification decision. 

\vspace{-10pt}
\subsection{Diagnostic Experiment}\label{sec:ablation}
\vspace{-4pt}

To$_{\!}$ perform$_{\!}$ extensive$_{\!}$ ablation experiments,$_{\!}$ we$_{\!}$~train ResNet101$_{\!}$ classification$_{\!}$ and$_{\!}$ DeepLab$_\text{V3}$$_{\!}$
seg- mentation$_{\!}$ models$_{\!}$ for$_{\!}$ $100$$_{\!}$ epochs$_{\!}$ and$_{\!}$ $80$K$_{\!}$ iterations, on$_{\!}$ ImageNet$_{\!}$ and$_{\!}$ ADE20K, respectively.

\vspace{-4pt}
\noindent\textbf{Class Sub-centroids.$_{\!\!}$} Table$_{\!}$~\ref{table:protocls} studies the impact of the
number of class sub-centroids ($K$) for each class. When $K\!=\!1$, each class is represented by its centroid -- the average feature vector of all the training samples of the class (Eq.~\ref{eq:DNC1}), without clustering. The corresponding baseline obtains $77.31\%$ \texttt{top-1} acc. and $43.2\%$ \texttt{mIoU} for classification and~segmentation, respectively. For classification, increasing $K$ from 1 to 4 leads to~better performance$_{\!}$ (\ie, $77.31\%\!\rightarrow\!77.80\%$).$_{\!}$ This$_{\!}$ supports$_{\!}$ our hypothesis$_{\!}$ that$_{\!}$ one$_{\!}$ single$_{\!}$ class$_{\!}$ weight/center$_{\!}$ is$_{\!}$ far$_{\!}$ from$_{\!}$ enough$_{\!}$ to$_{\!}$ capture$_{\!}$ the$_{\!}$ underlying$_{\!}$ data$_{\!}$ distribution and proves the efficacy of our clustering based sub-class pattern mining. We stop using $K_{\!}>_{\!}4$ as the required memory exceeds the computational limit of our hardware. We find
 similar$_{\!}$ trends$_{\!}$ on$_{\!}$ segmentation;$_{\!}$ using$_{\!}$ more$_{\!}$ sub-centroids$_{\!}$ ($K_{\!\!\!}:_{\!\!} 1_{\!\!}\rightarrow_{\!\!}10$)$_{\!\!}$ brings$_{\!}$~noticeable$_{\!}$ performance$_{\!}$ boost:$_{\!}$ 43.2\%$\rightarrow$ 44.3\%.$_{\!}$ However,$_{\!}$ increasing$_{\!}$ $K$$_{\!}$ above$_{\!}$ 10$_{\!}$ provides marginal$_{\!}$ or$_{\!}$ even$_{\!}$ negative$_{\!}$ gain.$_{\!}$ This$_{\!}$ may$_{\!}$ be$_{\!}$ because$_{\!}$ over-clustering finds some insignificant patterns, which are trivial or harmful for decision-making.

\vspace{-4pt}
\noindent\textbf{External Memory.} We next study the influence of the external memory, only used in image classification. As shown in Table$_{\!}$~\ref{table:memory}, DNC gradually improves the performance as the increase of the memory size. It reaches $77.80\%$ \texttt{top-1} acc. at size $1000$. However, the results are still not reaching the performance saturating point, but rather the upper limit of our hardware's computational budget.

\begin{table}[t]
	\caption{A  set  of  ablative  experiments  on  ImageNet$_{\!}$~\cite{ImageNet}  \texttt{val} and ADE20K$_{\!}$~\cite{zhou2017scene}$_{\!}$~\texttt{val}.}
\vspace{-7px}
		\hspace{-0.7em}
\begin{subtable}{0.36\linewidth}
\captionsetup{width=.80\linewidth}
		\resizebox{\textwidth}{!}{
			\setlength\tabcolsep{4pt}
					\renewcommand\arraystretch{1.0}
			\begin{tabular}{|c||cc|c||c|}
				\hline\thickhline
				\rowcolor{mygray}
&\multicolumn{2}{c|}{ImageNet} &&ADE20K\\
\rowcolor{mygray}
				\multirow{-2}{*}{$K$} & \texttt{top-1} & \texttt{top-5} &\multirow{-2}{*}{$K$} &\texttt{mIoU}\\
				
				\hline\hline
				1   & 77.31$\%$ &93.01$\%$&1 & 43.2$\%$\\
				2   & 77.54$\%$ &93.32$\%$&5 & 44.0$\%$\\
				3   & 77.68$\%$ &93.63$\%$&\textbf{10}& \textbf{44.3}$\%$\\
				\textbf{4}   & \textbf{77.80}$\%$ &\textbf{93.85}$\%$&20& 44.0$\%$\\
				\hline
			\end{tabular}
		}
		\vspace{4px}
		\setlength{\abovecaptionskip}{0.3cm}
		\setlength{\belowcaptionskip}{-0.1cm}
		\caption{Number$_{\!}$ of$_{\!}$ sub-centroids$_{\!}$ {\color{white}~~~~~~~~~~} ($K$) for each class}
		\vspace{-7px}
		\label{table:protocls}
	\end{subtable}
	\begin{subtable}{0.275\linewidth}
		\resizebox{\textwidth}{!}{
			\setlength\tabcolsep{4.5pt}
			\begin{tabular}{|c||cc|}
				\hline\thickhline
				\rowcolor{mygray}
				\rowcolor{mygray}
                Memory&\multicolumn{2}{c|}{ImageNet}\\
                \rowcolor{mygray}
				$_{\!\!}$(\#\textit{batch})$_{\!\!}$ & \texttt{top-1} & \texttt{top-5} \\
				\hline
				\hline
				0   & 77.49$\%$ &93.09$\%$ \\
				700   & 77.58$\%$ &93.16$\%$ \\
				800   & 77.64$\%$ &93.35$\%$ \\
				900   & 77.75$\%$ &93.67$\%$ \\
				\textbf{1000}  & \textbf{77.80}$\%$ &\textbf{93.85}$\%$ \\
				\hline
			\end{tabular}
		}
		\vspace{-0px}
		\setlength{\abovecaptionskip}{0.4cm}
		\setlength{\belowcaptionskip}{-0.1cm}
		\caption{Memory size}
		\vspace{-6px}
		\label{table:memory}
		\end{subtable}
	\begin{subtable}{0.355\linewidth}
		\resizebox{\textwidth}{!}{
			\setlength\tabcolsep{4.5pt}
			\begin{tabular}{|c||cc|c|}
				\hline\thickhline
				\rowcolor{mygray}
                &\multicolumn{2}{c|}{ImageNet} &ADE20K\\
                \rowcolor{mygray}
				\multirow{-2}{*}{$\mu$} & \texttt{top-1} & \texttt{top-5} &\texttt{mIoU}\\
				\hline
				\hline
				$0$ & 73.82$\%$ & 93.02$\%$ & 42.7$\%$ \\
				$0.9$ & 76.41$\%$ & 93.07$\%$ & 43.6$\%$\\
				$0.99$ & 77.33$\%$ & 93.51$\%$ & 44.0$\%$\\
				$\textbf{0.999}$ & \textbf{77.80}$\%$ &\textbf{93.85}$\%$ & \textbf{44.3}$\%$\\
				$0.9999$ &  77.31$\%$ & 93.48$\%$ & 44.2$\%$ \\
				\hline
			\end{tabular}
		}
		\vspace{-0px}
		\setlength{\abovecaptionskip}{0.4cm}
		\setlength{\belowcaptionskip}{-0.1cm}
		\caption{Momentum  coefficient ($\mu$)}
		\vspace{-6px}
		\label{table:muc}
	\end{subtable}
\vspace{-18pt}
\end{table}

\vspace{-4pt}
\noindent\textbf{Momentum Update.} Last, we ablate the effect of the momentum coefficient $\mu$ (Eq.~\ref{eq:mom}) that controls the speed of sub-centroid online updating. From Table~\ref{table:muc} we find the behaviors of $\mu$ are consistent in both tasks. In particular, DNC performs well with larger coefficients (\ie, $\mu\!\in\![0.999, 0.9999]$), signifying the importance of slow updating. The performance degrades when $\mu\!\in\![0.9, 0.99]$, and encounters a large drop when $\mu\!=\!0$ (\ie, only using batch sub-centroids as approximations).



\vspace{-10pt}
\section{Conclusion}
\vspace{-8pt}
We present deep$_{\!}$ nearest$_{\!}$ centroids$_{\!}$ (DNC), building upon the classic idea of classifying data samples according to nonparametric class~representatives. Compared to classic parametric$_{\!}$ models,$_{\!}$ DNC$_{\!}$ has$_{\!}$ merits$_{\!}$ in:$_{\!}$ \textbf{i)}$_{\!}$ systemic$_{\!}$ simplicity$_{\!}$~by$_{\!}$ bringing the intuitive Nearest Centroids mechanism to DNNs; \textbf{ii)} automated discovery of latent data structure using within-class clustering; \textbf{iii)} direct supervision~of representation$_{\!}$ learning,$_{\!}$ boosted$_{\!}$ by$_{\!}$ unsupervised$_{\!}$ sub-pattern$_{\!}$ mining;$_{\!}$ \textbf{iv)}$_{\!}$ improved$_{\!}$ transferability$_{\!}$ that lossless$_{\!}$ transfers$_{\!}$ learnable$_{\!}$ knowledge$_{\!}$  across$_{\!}$ tasks;$_{\!}$ and$_{\!}$ \textbf{v)}$_{\!}$ \textit{ad-hoc}$_{\!}$ explainability$_{\!}$ by$_{\!}$ anchoring$_{\!}$ class$_{\!}$ exemplars$_{\!}$ with$_{\!}$ real$_{\!}$ observations. Experiments confirm the efficacy and enhanced interpretability.

\section*{Acknowledgement}

This research was supported by the National Science Foundation under Grant No. 2242243.

\bibliography{iclr2023_conference}
\bibliographystyle{splncs04}

\clearpage
\appendix
\centerline{\maketitle{\textbf{SUMMARY OF THE APPENDIX}}}

This appendix contains additional details for the ICLR 2023 submission, titled \textit{Visual Recognition with Deep Nearest Centroids"}. The appendix is organized as follows:
\begin{itemize}
  \item \S\ref{sec_app:A1} provides the pseudo code of DNC.
  \item \S\ref{sec_app:A2} introduces more quantitative results on image classification.
  \item \S\ref{sec_app:A3} gathers additional semantic segmentation results on COCO-Stuff \cite{caesar2018coco} dataset.
  \item \S\ref{sec_app:A4} presents the corresponding error bars on Tables$_{\!}$~\ref{table:cifar10}, \ref{table:imagenet}, and \ref{table:seg}.
  \item \S\ref{sec_app:A5} investigates the potential of DNC in sub-categories discovery.
  \item \S\ref{sec_app:A5_add_1} reports the transferability of DNC towards other image classification task.
    \item \S\ref{sec_app:A5_add_2} evaluates the performance of DNC on ImageNetv2 \texttt{test} sets.
   \item \S\ref{sec_app:A5_add_3} compares the performance of using $k$-means and Sinkhorn-Knopp clustering algorithms.
\item \S\ref{sec_app:A5_add_4} compares DNC with different distance (learning) based classifiers.
  \item \S\ref{sec_app:A6} reports additional diagnostic experiments for further investigations on $K$, update ratio of external memory, ImageNet capacity, feature size, and temperature parameter $\varepsilon$ in (\ref{eq:BIP2}).
  \item \S\ref{sec_app:A7} offers more detailed discussions regarding the GPU memory cost.
  \item \S\ref{sec_app:A8} depicts more visual examples regarding the \textit{ad-hoc} explainability of DNC.
  \item \S\ref{sec_app:A9} plots qualitative semantic segmentation results.
  \item \S\ref{sec_app:B} gives additional review of representative literature on metric-/distance-learning and clustering-based unsupervised representation learning.
  \item \S\ref{sec_app:C} discusses our limitations, societal impact, and directions of our future work.
\end{itemize}

\clearpage
\vspace{-6pt}
\section{Pseudo Code of DNC and Code Release}
\label{sec_app:A1}
\vspace{-3pt}
The pseudo-code of DNC is given in Algorithm \ref{alg:code}. To guarantee reproducibility, our code is available at \href{https://github.com/ChengHan111/DNC}{https://github.com/ChengHan111/DNC}. 
\vspace{8pt}


\begin{algorithm}[H]
    \caption{Pseudo-code of DNC in a PyTorch-like style.}
    \label{alg:code}
    \definecolor{codeblue}{rgb}{0.25,0.5,0.5}
    \lstset{
     backgroundcolor=\color{white},
     basicstyle=\fontsize{9pt}{9pt}\ttfamily\selectfont,
     columns=fullflexible,
     breaklines=true,
     captionpos=b,
     commentstyle=\fontsize{9pt}{9pt}\color{codeblue},
     keywordstyle=\fontsize{9pt}{9pt},
    }
   \begin{lstlisting}[language=python]
   # P: non-parametric sub-centroids (C x K x D)
   # X: feature embeddings (N x D)

   # C: number of classes
   # K: number of sub-centroids for each class
   # R: sinhorn-knopp iteration number
   # mu: momentum coefficient (Eq.8)
   # epsilon: hyper-parameter in the Sinkhorn-Knopp algorithm (Eq.6)

   def DNC(X, label)
       #== Model Prediction and Training Loss (Eq.7) ==#

       # image-to-centroid assignment (N x K x C, Eq.5)
       L = torch.einsum('nd,ckd->nkc', X, P)
       output = torch.amax(L, dim=1)
       loss = CrossEntropyLoss(output, label)

       #======= Sub-centroid Estimation =======#
       for c in range(C)
           init_L = L[...,c]
           Q = online_clustering(init_L)

           # assignments and embeddings for images in class c
           m_c = L[label == c]
           x_c = X[label == c, ...]

           # find images that are assigned to each sub-centroid
           # and correctly classified
           m_c_tile = repeat(m_c, tile=K)
           m_q = Q * m_c_tile

           # find images with label c that are correctly classified
           x_c_tile = repeat(m_c, tile=x_c.shape[-1])
           x_c_q = x_c * x_c_tile
           f = torch.mm(m_q.transpose(), x_c_q)

           # num assignments for each sub-centroid of class c
           n = torch.sum(m_q, dim=0)

           # momentum update (Eq.8)
           if torch.sum(n) > 0:
               P_c = mu * P[c, n != 0, :] + (1-mu) * f[n != 0, :]
               P[c, n != 0, :] = P_c

       return loss

   def online_clustering(L, iters=3, epsilon=0.05)
       Q = torch.exp(L / epsilon)
       Q /= torch.sum(Q)

       for _ in range(R):
          # row normalization
          Q /= torch.sum(Q, dim=1, keepdim=True)
          Q /= K

          # column normalization
          Q /= torch.sum(L, dim=0, keepdim=True)
          Q /= N

       # make sure the sum of each column to be 1
       Q *= N

       return one_hot(Q)
   \end{lstlisting}
   \end{algorithm}
\clearpage
\section{More Experiments on Image Classification}\label{sec_app:A2}

\textbf{Additional Results on CIFAR-10 \cite{CIFAR}.} CIFAR-10 dataset contains $60$K ($50$K/$10$K for \texttt{train}/ \texttt{test}) $32\!\times\!32$ colored images of $10$ classes.  Table$_{\!}$~\ref{table:cifar10_supp} reports additional comparison results on CIFAR-10, based on ResNet-18$_{\!}$~\cite{he2016deep} network architecture.$_{\!}$ As$_{\!}$ seen,$_{\!}$ in terms of \texttt{top-1} accuracy, our$_{\!}$  DNC$_{\!}$ is$_{\!}$ \textbf{0.94}\%$_{\!}$
higher$_{\!}$ than$_{\!}$ the$_{\!}$ parametric$_{\!}$ counterpart,$_{\!}$ under$_{\!}$ the$_{\!}$ same$_{\!}$ training$_{\!}$ setting.

\setlength\intextsep{10pt}
\begin{table}[H]
\centering
\captionsetup{width=.57\textwidth}
	\setlength{\abovecaptionskip}{0cm}
	\caption{\textbf{Classification \texttt{top-1} accuracy} on CIFAR-10~\cite{CIFAR} \texttt{test}.
		{\#Params: the number of learnable parameters (same for other tables). See \S\ref{sec_app:A2} for more details.}}
\vspace{3pt}
\hspace{-0.6em}
	\resizebox{0.55\textwidth}{!}{
		\setlength\tabcolsep{4pt}
		\renewcommand\arraystretch{1.1}
		\begin{tabular}{|rc||cc| }
			\hline\thickhline
			\rowcolor{mygray}
			Method~~~~ & Backbone & \#Params& \texttt{top-1} \\	\hline	\hline 
			ResNet \cite{he2016deep} &\multirow{2}{*}{ResNet18} & 11.17M & 93.55\% \\
            \textbf{DNC}-ResNet & & 11.16M &\textbf{94.49}\%\\\hline
	\end{tabular}}
	\label{table:cifar10_supp}
\end{table}

\noindent\textbf{Additional Results on CIFAR-100 \cite{CIFAR}.} Table$_{\!}$~\ref{table:cifar100_supp} reports comparison results on CIFAR-100, based$_{\!}$ on$_{\!}$~ResNet50 and ResNet101 network architectures$_{\!}$~\cite{he2016deep}. CIFAR-100 dataset has 100 classes with 500 training images and 100 testing images per class. We can find that, DNC obtains consistently better performance, compared to the classic parametric counterpart. Specifically,  DNC is \textbf{0.10}\% higher on ResNet50, and \textbf{0.16}\% higher on ResNet101, in terms of \texttt{top-1} accuracy.

\setlength\intextsep{10pt}
\begin{table}[H]
\centering
\captionsetup{width=.57\textwidth}
	\setlength{\abovecaptionskip}{0cm}
	\caption{\textbf{$_{\!}$Classification$_{\!}$ \texttt{top-1}$_{\!}$ accuracy$_{\!}$} on$_{\!}$ CIFAR-100$_{\!}$~\cite{CIFAR}$_{\!}$ \texttt{test}, using lightweight backbone network architectures : MobileNet-V2~\cite{sandler2018mobilenetv2}, and Swin-T$_{\!}$~\cite{liu2021swin}. See \S\ref{sec_app:A2} for more details.}
\vspace{5pt}
\hspace{-0.8em}
	\resizebox{0.56\textwidth}{!}{
		\setlength\tabcolsep{4pt}
		\renewcommand\arraystretch{1.1}
		\begin{tabular}{|rc||cc| }
			\hline\thickhline
			\rowcolor{mygray}
			Method~~~~ & Backbone & \#Params& \texttt{top-1} \\	\hline	\hline 
			ResNet \cite{he2016deep} &\multirow{2}{*}{ResNet50} & 23.71M & 79.81\% \\
            \textbf{DNC}-ResNet & & 23.50M &\textbf{79.91}\%\\\hline
			ResNet \cite{he2016deep} &\multirow{2}{*}{ResNet101} & 42.70M & 79.83\%\\  
            \textbf{DNC}-ResNet & &42.49M & \textbf{79.99}\%\\
			\hline
	\end{tabular}}
	\label{table:cifar100_supp}
\end{table}

\noindent\textbf{Additional$_{\!}$ Results$_{\!}$ on$_{\!}$ ImageNet$_{\!}$~\cite{ImageNet}$_{\!}$~using$_{\!}$ Lightweight$_{\!}$ Backbone$_{\!}$ Network$_{\!}$ Architectures.}$_{\!}$~Table~\ref{table:imagenet_lightweight}$_{\!}$ reports performance on ImageNet$_{\!}$~\cite{ImageNet},$_{\!}$ using$_{\!}$ two$_{\!}$ lightweight$_{\!}$ backbone architectures:  MobileNet-V2~\cite{sandler2018mobilenetv2}, and Swin-T$_{\!}$~\cite{liu2021swin}. As can be seen, DNC, again, attributes decent performance. In particular, our DNC is \textbf{0.28}\% and \textbf{0.30}\% higher on MobileNet-V2 and Swin-Tiny, respectively. It is worth noticing that DNC can efficiently reduce the number of learnable parameters when the parametric classifier occupies a massive proportion in the original lightweight classification networks. Taking MobileNet-V2 as an example: our DNC reduces the number of learnable parameters from 3.50M to \textbf{2.22}M. 

\begin{table}[H]
\centering
\captionsetup{width=.58\textwidth}
	\setlength{\abovecaptionskip}{0cm}
	\caption{\textbf{$_{\!}$Classification$_{\!}$ \texttt{top-1}$_{\!}$ accuracy$_{\!}$} on$_{\!}$ ImageNet$_{\!}$~\cite{ImageNet}$_{\!}$ \texttt{val}. See \S\ref{sec_app:A2} for more details.}
\vspace{5pt}
\hspace{-0.8em}
	\resizebox{0.58\textwidth}{!}{
		\setlength\tabcolsep{4pt}
		\renewcommand\arraystretch{1.1}
		\begin{tabular}{|rc||cc| }
			\hline\thickhline
			\rowcolor{mygray}
			Method~~~~ & Backbone & \#Params& \texttt{top-1} \\	\hline	\hline 
            MobileNet \cite{sandler2018mobilenetv2} &\multirow{2}{*}{MobileNet-V2} & 3.50M & 71.76\%\\  
            \textbf{DNC}-MobileNet & &\textbf{2.22M} & \textbf{72.04}\%\\
            \hline
			Swin \cite{liu2021swin} &\multirow{2}{*}{Swin-T} & 28.29M & 81.08\%\\  
            \textbf{DNC}-Swin & &\textbf{27.52M} & \textbf{81.38}\%\\
			\hline
	\end{tabular}}
	\label{table:imagenet_lightweight}
\end{table}

\clearpage
\section{More Experiments on Semantic Segmentation}
\label{sec_app:A3}

\vspace{-0.1cm}
For through evaluation, we conduct extra experiments on COCO-Stuff \cite{caesar2018coco}, a famous semantic segmentation dataset. COCO-Stuff contains $9$K/$1$K images for \texttt{train/test} of 80 object classes and 91 stuff classes. Similar to \S\ref{sec:seg}, we approach DNC on FCN$_{\!}$~\cite{long2015fully},$_{\!}$ DeepLab$_\text{V3}$$_{\!\!}$~\cite{chen2017rethinking},$_{\!}$ and UperNet$_{\!}$~\cite{xiao2018unified},$_{\!}$ using$_{\!}$ two$_{\!}$ backbone architectures, \ie, ResNet101$_{\!}$~\cite{he2016deep} and Swin-B$_{\!}$~\cite{liu2021swin}.  All models are obtained by following the standard training settings of COCO-Stuff \texttt{train}, \ie, crop size $512\!\times\!512$, batch size $16$, and $40$K iterations. From Table \ref{table:seg_coco} we can draw similar conclusions: in comparison with the parametric counterpart, DNC produces more precise segments and yields improved transferability.

\vspace{-0.1cm}

\begin{table}[H]
	\centering
\captionsetup{width=.65\textwidth}
	\caption{{\textbf{Segmentation mIoU score} on COCO-Stuff \cite{caesar2018coco}. See \S\ref{sec_app:A3} for more details.}}
\hspace{-0.6em}
	\resizebox{0.62\columnwidth}{!}{
		\setlength\tabcolsep{3.5pt}
		\renewcommand\arraystretch{1.1}
		\begin{tabular}{|r|c||c| }
			\hline\thickhline
			\rowcolor{mygray}
			& & COCO-Stuff \texttt{test} \\
            \rowcolor{mygray}
            \multirow{-2}{*}{Method~~~~} & \multirow{-2}{*}{Backbone} &  \texttt{mIoU} \\	\hline	\hline
			\multirow{2}{*}{FCN~\cite{long2015fully}} &ResNet101~\cite{he2016deep} & 32.63\% \\
             &\textbf{DNC}-ResNet101 & \reshx{32.89\%}{0.26} \\
            \multirow{2}{*}{\textbf{DNC}-FCN} &ResNet101~\cite{he2016deep}& \reshx{33.04\%}{0.41} \\
             &\textbf{DNC}-ResNet101  & \reshx{\textbf{33.49}\%}{0.86} \\\hline

			\multirow{2}{*}{DeepLab$_\text{V3}$~\cite{chen2017rethinking}} &ResNet101~\cite{he2016deep} & 36.01\% \\
             &\textbf{DNC}-ResNet101 & \reshx{36.28\%}{0.27} \\
            \multirow{2}{*}{\textbf{DNC}-DeepLab$_\text{V3}$} &ResNet101~\cite{he2016deep}& \reshx{36.51\%}{0.50} \\
             &\textbf{DNC}-ResNet101  & \reshx{\textbf{36.79}\%}{0.78} \\\hline

            \multirow{2}{*}{UperNet~\cite{xiao2018unified}} &Swin-B~\cite{liu2021swin} & 42.77\% \\
             &\textbf{DNC}-Swin-B & \reshx{42.84\%}{0.07} \\
            \multirow{2}{*}{\textbf{DNC}-UperNet} &Swin-B~\cite{liu2021swin}& \reshx{43.13\%}{0.36} \\
             &\textbf{DNC}-Swin-B  & \reshx{\textbf{43.29}\%}{0.52} \\\hline
	\end{tabular}
	}
\vspace{-8pt}
	\label{table:seg_coco}
\end{table}

\vspace{-0.4cm}
\section{Error Bars}
\label{sec_app:A4}
\vspace{-0.4cm}
In this section, we report standard deviation error bars on Tables$_{\!}$~\ref{table:cifar10_err}, \ref{table:imagenet_err}, and \ref{table:seg_err} for our main experiments regarding image classification (\S\ref{sec:cls}) and semantic segmentation (\S\ref{sec:seg}). The results are obtained by training the algorithm three times, with different initialization seeds.

\vspace{-0.1cm}
\begin{table}[H]
\centering
	\setlength{\abovecaptionskip}{0cm}
\captionsetup{width=.65\textwidth}
	\caption{\textbf{Classification \texttt{top-1} accuracy} on CIFAR-10$_{\!}$~\cite{CIFAR} \texttt{test} with error bars.  See \S\ref{sec_app:A4} for more details.}
\vspace{5pt}
\hspace{-0.6em}
	\resizebox{0.65\textwidth}{!}{
		\setlength\tabcolsep{4pt}
		\renewcommand\arraystretch{1.1}
		\begin{tabular}{|rc||cc| }
			\hline\thickhline
			\rowcolor{mygray}
			Method~~~~ & Backbone & \#Params& \texttt{top-1} \\	\hline	\hline 
			ResNet \cite{he2016deep} &\multirow{2}{*}{ResNet50} & 23.52M & 95.55 $\pm$ (0.14)\% \\
            \textbf{DNC}-ResNet & & 23.50M &\textbf{95.78} $\pm$ (0.12)\%\\\hline
			ResNet \cite{he2016deep} &\multirow{2}{*}{ResNet101} & 42.51M & 95.58 $\pm$ (0.13)\%\\  
            \textbf{DNC}-ResNet & &42.49M & \textbf{95.82} $\pm$ (0.13)\%\\
			\hline
	\end{tabular}}
	\label{table:cifar10_err}
\end{table}


\begin{table}[H]
\centering
	\setlength{\abovecaptionskip}{0cm}
	\setlength{\belowcaptionskip}{-0.2cm}
\captionsetup{width=.75\textwidth}
	\caption{\textbf{Classification$_{\!}$ \texttt{top-1}$_{\!}$ and \texttt{top-5}$_{\!}$ accuracy} on ImageNet~\cite{ImageNet} \texttt{val} with error bars. See \S\ref{sec_app:A4} for more details.}
\vspace{5pt}
\hspace{-0.6em}
	\resizebox{0.75\textwidth}{!}{
		\setlength\tabcolsep{6pt}
		\renewcommand\arraystretch{1.1}
		\begin{tabular}{|rc||ccc|}
			\hline\thickhline
			\rowcolor{mygray}
			Method~~~~ & Backbone & \#Params & \texttt{top-1} & \texttt{top-5}  \\	\hline	\hline
			ResNet \cite{he2016deep} &\multirow{2}{*}{ResNet50} & 25.56M & 76.20 $\pm$ (0.10)\%& 93.01\%\\
            \textbf{DNC}-ResNet& &23.51M& \textbf{76.49} $\pm$ (0.09)\% & \textbf{93.08}\%\\\hline
			ResNet \cite{he2016deep} &\multirow{2}{*}{ResNet101} & 44.55M & 77.52 $\pm$ (0.11)\%& 93.06\%\\
			\textbf{DNC}-ResNet& &42.50M& \textbf{77.80} $\pm$ (0.10)\% & \textbf{93.85}\% \\\hline
            Swin \cite{liu2021swin} &\multirow{2}{*}{Swin-S} & 49.61M & 83.02 $\pm$ (0.14)\%& 96.29\%\\
            \textbf{DNC}-Swin& &48.84M& \textbf{83.26} $\pm$ (0.13)\%& \textbf{96.40}\% \\\hline
            Swin \cite{liu2021swin} &\multirow{2}{*}{Swin-B} & 87.77M & 83.36 $\pm$ (0.12)\%& 96.44\%\\
            \textbf{DNC}-Swin& &86.75M& \textbf{83.68} $\pm$ (0.12)\%& \textbf{97.02}\%\\ \hline
	\end{tabular}}
	\label{table:imagenet_err}
\end{table}


\begin{table}[H]
\centering
	\setlength{\abovecaptionskip}{0cm}
	\setlength{\belowcaptionskip}{-0.2cm}
\captionsetup{width=.9\textwidth}
		\caption{\textbf{Segmentation mIoU score} on ADE20K~\cite{zhou2017scene} \texttt{val} and Cityscapes \cite{cordts2016cityscapes} \texttt{val} with error bars. See \S\ref{sec_app:A4} for more details.}
\vspace{5pt}
	\resizebox{0.98\columnwidth}{!}{
		\setlength\tabcolsep{4pt}
		\renewcommand\arraystretch{1.05}
		\begin{tabular}{|r|cc||ccc|}
			\hline\thickhline
			\rowcolor{mygray}
			&   &${\!\!\!\!\!\!}$ImageNet  & &{ADE20K} & {Cityscapes} \\
			\rowcolor{mygray}
			\multirow{-2}{*}{Method~~~~~~}& \multirow{-2}{*}{Backbone} &{\texttt{top-1} acc.}  & \multirow{-2}{*}{\#Params} & \texttt{mIoU}& \texttt{mIoU} \\ \hline\hline
				
			&ResNet101~\cite{he2016deep}
			& 77.52\% & 68.6M &39.9 $\pm$ (0.11)\% & 75.6 $\pm$ (0.13)\% \\
						
			\multirow{-2}{*}{FCN~\cite{long2015fully}}
			&\textbf{DNC}-ResNet101 &77.80\% &68.6M & 40.4 $\pm$ (0.11)\%& 76.3 $\pm$ (0.12)\% \\

			&ResNet101~\cite{he2016deep}
			& 77.52\%   &68.5M & 41.1 $\pm$ (0.10)\% & 76.7 $\pm$ (0.11)\% \\
			
			\multirow{-2}{*}{\textbf{DNC}-FCN} &\textbf{DNC}-ResNet101
			& 77.80\% &68.5M & \textbf{42.3} $\pm$ (0.10)\%   & \textbf{77.5} $\pm$ (0.11)\% \\ \hline

			&ResNet101~\cite{he2016deep}
			& 77.52\% & 62.7M &44.1 $\pm$ (0.13)\% & 78.1 $\pm$ (0.12)\%\\
			
			\multirow{-2}{*}{DeepLab$_\text{V3}$~\cite{chen2017rethinking}}
			&\textbf{DNC}-ResNet101  &77.80\% &62.7M & 44.6 $\pm$ (0.13)\% & 78.7 $\pm$ (0.13)\% \\

			&ResNet101~\cite{he2016deep}
			& 77.52\% &62.6M & 45.0 $\pm$ (0.10)\%  & 79.1 $\pm$ (0.13)\% \\
			
			\multirow{-2}{*}{\textbf{DNC}-DeepLab$_\text{V3}$}
			&\textbf{DNC}-ResNet101 & 77.80\% &62.6M & \textbf{45.7} $\pm$ (0.09)\% & \textbf{79.8} $\pm$ (0.12)\% \\ \hline

			&Swin-B~\cite{liu2021swin}
			& 83.36\% &90.6M & 48.0 $\pm$ (0.11)\% & 79.8 $\pm$ (0.13)\% \\

			\multirow{-2}{*}{UperNet~\cite{xiao2018unified}}
			&\textbf{DNC}-Swin-B &83.68\%  &90.6M  & 48.4  $\pm$ (0.10) \%  & 80.1 $\pm$ (0.12)\% \\

			&Swin-B~\cite{liu2021swin}
			&83.36\% &90.5M & 48.6 $\pm$ (0.09) \% & 80.5 $\pm$ (0.10)\% \\
			
			\multirow{-2}{*}{\textbf{DNC}-UperNet}&\textbf{DNC}-Swin-B
			&83.68\% &90.5M & \textbf{50.5} $\pm$ (0.09)\% & \textbf{80.9} $\pm$ (0.10)\% \\ \hline
	\end{tabular}}
	\label{table:seg_err}
\end{table}

\vspace{-0.4cm}
\section{Experiments on Sub-Categories Discovery}
\label{sec_app:A5}

Through unsupervised, within-class clustering, our DNC represents each class as a set of automatically discovered class sub-centroids (\ie, cluster center). This allows DNC to better describe the underlying, multimodal data structure and robusty depict for rich intra-class variance. In other words, our DNC can effectively capture sub-class patterns, which is conducive to algorithmic performance. Such a capacity of sub-patter mining is also considered crucial for good transferable features -- representations learnt on coarse classes are capable of fine-grained recognition~\cite{huh2016makes}.


In order to quantify the ability of DNC for automatic sub-category discovery, we follow the experimental setup posed by~\cite{huh2016makes} -- learning the feature embedding using coarse-grained
object labels, and evaluating the learned feature using fine-grained object labels. This evaluation strategy allows us to assess the feature learning performance regarding how well the deep model can discover variations within each category. A conjecture is that deep networks that perform well on this test have an exceptional capacity to identify and mine sub-class patterns during training, which the proposed DNC seeks to rigorously establish.

In particular, the network is first trained on coarse-grained labels with the baseline parametric softmax and with our non-parametric DNC using the same network architecture. After training on coarse classes,  we use the top-1 nearest neighbor accuracy in the final feature space to measure the accuracy of identifying fine-grained classes. The classification performance evaluated in such setting is referred as \textbf{induction accuracy} as in \cite{huh2016makes}.
Next we provide our experimental results on CIFAR100~\cite{CIFAR} and ImageNet~\cite{ImageNet}, respectively.

\noindent\textbf{Performance of Sub-category Discovery on CIFAR100.} CIFAR100 includes both fine-grained annotations in 100 classes and coarse-grained annotations in 20 classes. We examine sub-category discovery by transferring representation learned from 20 classes to 100 classes. As shown in Table~\ref{table:knn_cifar100}, DNC consistently outperforms the parametric counterpart: DNC increases 0.12\%, in terms of the standard \texttt{top-1} accuracy, on both ResNet50 and ResNet101 architectures. Nevertheless, when transferred to CIFAR100 (\ie, 100 classes) using $k$-NN$_{\!}$, a significant loss occurs on the baseline: 53.22\% and 54.31\% \texttt{top-1}$_{\!}$~acc. on ResNet50 and ResNet101, respectively. Our features, on the other hand, provide 14.24\% and 13.82\% improvement over the baseline, achieving 67.46\% and 68.13\% \texttt{top-1}$_{\!}$~acc. on 100 classes on ResNet50 and ResNet101, respectively. In addition, in comparison the parametric model, our approach results in only a smaller drop in transfer performance, \ie, \textbf{-18.87} \textit{vs} -32.99 on ResNet50, and \textbf{-18.47} \textit{vs} -32.17 on ResNet101.

\begin{table}[H]
	\centering
\captionsetup{width=.65\textwidth}
	\caption{\textbf{\texttt{Top-1}$_{\!}$ induction accuracy$_{\!}$} on$_{\!}$ CIFAR-100~\cite{CIFAR}$_{\!}$ \texttt{test} using CIFAR-20 pre-trained models. Numbers reported with $k$-nearest neighbor classifiers. See \S\ref{sec_app:A5} for more details.}
\hspace{-0.6em}
	\resizebox{0.63\columnwidth}{!}{
		\setlength\tabcolsep{3.5pt}
		\renewcommand\arraystretch{1.1}
		\begin{tabular}{|rc||cc| }
			\hline\thickhline
			\rowcolor{mygray}
			Method~~~~ & Backbone & 20~classes & 100~classes \\	\hline	\hline 
			ResNet \cite{he2016deep} &\multirow{2}{*}{ResNet50} & 86.21\% & 53.22\% \\
            \textbf{DNC}-ResNet & & \textbf{86.33}\% &\textbf{67.46}\%\\\hline
			ResNet \cite{he2016deep} &\multirow{2}{*}{ResNet101} & 86.48\% & 54.31\%\\  
            \textbf{DNC}-ResNet & &\textbf{86.60}\% & \textbf{68.13}\%\\
			\hline
	\end{tabular}
	}
	\label{table:knn_cifar100}
\end{table}

\vspace{-0.4cm}
{\noindent\textbf{Performance of Sub-category Discovery on ImageNet.$_{\!}$} Table~\ref{table:knn_imagenet} provides experimental results of sub-category discovery on ImageNet \texttt{val}. As in \cite{huh2016makes}, 127 coarse ImageNet categories are obtained by top-down clustering of 1K ImageNet categories on WordNet tree. Training on the 127 coarse classes, DNC improves the performance of baseline by 0.10\% and 0.03\%, achieving 84.39\% and 85.91\% on ResNet50 and ResNet101, respectively. When transferring to the 1K ImageNet classes using $k$-NN$_{\!}$, our features provide huge improvements, \ie, 8.98\% and 9.29\%,  over the baseline.}

\begin{table}[H]
	\centering
\captionsetup{width=.65\textwidth}
	\caption{\textbf{\texttt{Top-1}$_{\!}$ induction accuracy$_{\!}$} on$_{\!}$ ImageNet~\cite{ImageNet} \texttt{val} using ImageNet-127 pre-trained models. Numbers reported with $k$-nearest neighbor classifiers. See \S\ref{sec_app:A5} for more details.}
\hspace{-0.6em}
	\resizebox{0.7\columnwidth}{!}{
		\setlength\tabcolsep{4.5pt}
		\renewcommand\arraystretch{1.1}
		\begin{tabular}{|rc||cc| }
			\hline\thickhline
			\rowcolor{mygray}
			Method~~~~ & Backbone & 127~classes & 1000~classes \\	\hline	\hline 
			ResNet \cite{he2016deep} &\multirow{2}{*}{ResNet50} & 84.29\% & 43.23\% \\
            \textbf{DNC}-ResNet & & 84.39\% &\textbf{52.21}\%\\\hline
			ResNet \cite{he2016deep} &\multirow{2}{*}{ResNet101} & 85.88\% & 45.31\%\\  
            \textbf{DNC}-ResNet & &85.91\% & \textbf{54.60}\%\\
			\hline
	\end{tabular}
	}
	\label{table:knn_imagenet}
\end{table}

The promising transfer results on CIFAR100 and ImageNet serve as strong evidence to suggest that our DNC is capable of automatically discovering meaningful sub-class patterns -- latent visual structures that are not explicitly presented in the supervisory signal, and hence handle intra-class variance and boost visual recognition.

\section{Transferability towards other image classification task}
\label{sec_app:A5_add_1}
In addition to conducting the coarse-to-fine transfer learning experiment (\S\ref{sec_app:A5}), we further evaluate the transfer learning performance by applying ImageNet-trained weight to Caltech-UCSD Birds-200-2011 (CUB-200-2011) dataset \cite{wah2011caltech}, following \cite{wang2021self,oh2016deep,plested2022deep}.

Specifically, CUB-200-2011 dataset comprises 11,788 bird photos arranged into 200 categories, with 5,994 for training and 5,794 for testing.  All the models use ResNet50 architecture~\cite{he2016deep}, and are trained for 100 epochs. SGD optimizer is adopted, where the learning rate is initialized as 0.01 and organized following a polynomial annealing policy. Standard data augmentation techniques are used, including flipping, cropping and normalizing. Experimental results are reported in Table~\ref{table:CUB_transfer}. As seen, DNC is +\textbf{0.73}\% and +\textbf{0.39}\% higher in \texttt{Top-1} and \texttt{Top-5} acc., respectively. This clearly verifies that DNC owns better transferability.

\begin{table}[H]
\centering
\captionsetup{width=.70\textwidth}
	\caption{\textbf{Classification$_{\!}$ \texttt{top-1}$_{\!}$ and \texttt{top-5}$_{\!}$ accuracy} on CUB-200-2011 \texttt{test}~\cite{wah2011caltech}.  See \S\ref{sec_app:A5_add_1} for more details.}
\hspace{-0.6em}
	\resizebox{0.7\textwidth}{!}{
		\setlength\tabcolsep{6pt}
		\renewcommand\arraystretch{1.1}
		\begin{tabular}{|rc||cc|}
			\hline\thickhline
			\rowcolor{mygray}
			Model (ImageNet-trained) & Backbone & \texttt{top-1} & \texttt{top-5}  \\	\hline	\hline
			ResNet \cite{he2016deep} &\multirow{2}{*}{ResNet50}  & 84.48\%& 96.31\%\\
            \textbf{DNC}-ResNet& & \textbf{85.21}\% & \textbf{96.70}\%\\\hline
	\end{tabular}}
	\label{table:CUB_transfer}
\end{table}

For parametric softmax classifier, both the feature network and softmax layer are fully learnable parameters. That means, for a new task, the softmax classifier has to both finetune the feature network and train a new softmax layer. However, for DNC, the feature network is the only learnable part. The class centers are not freely learnable parameters; they are directly computed from training data on the feature space. For a new task, DNC just needs to fine-tune the only learnable part --  the feature network; the class centers are still directly computed from the training data according to the clustering assignments, without end-to-end training. Hence DNC owns better transferability during network fine-tuning.

\section{Evaluation on ImageNetv2 test sets}
\label{sec_app:A5_add_2}

We evaluate DNC-ResNet50 on ImageNetv2 \cite{recht2019imagenet} test sets, \ie, ``Matched Frequency", ``Threshold0.7" and ``Top Images". Each test set contains 10 images for each ImageNet class, collected from MTurk. In particular,  each MTurk worker is assigned with a certain classes. Then each worker is asked to select images belonging to his/her target class, from several candidate images sampled from a large image pool as well as ImageNet validation set. The output is a \textit{selection frequency} for each image, \ie, the fraction of MTurk workers selected the image in a task for its target class. Then three test sets are developed according to different principles defined on the selection frequency.  For ``Matched Frequency", \cite{recht2019imagenet} first approximated the selection frequency distribution for each class using those ``re-annotated'' ImageNet validation images. According to these class-specific distributions, ten test images are sampled from the candidate pool for each class.  For ``Threshold0.7", \cite{recht2019imagenet} sampled ten images from each class with selection frequency at least 0.7. For ``Top Images", \cite{recht2019imagenet} selected the ten images with the highest selection frequency for each class. The results on these three test sets are shown in Table~\ref{table:ImageNetv2}. As seen, our DNC exceeds the parametric softmax based ResNet50 by +\textbf{0.52-0.89}\% \texttt{top-1} and +\textbf{0.26-0.47}\% \texttt{top-5} acc..

\begin{table}[H]
\centering
\captionsetup{width=.75\textwidth}
	\caption{\textbf{Classification \texttt{top-1}$_{\!}$ and \texttt{top-5}$_{\!}$ accuracy} on ImageNetv2 \texttt{test} sets~\cite{recht2019imagenet}.  See \S\ref{sec_app:A5_add_2} for more details.}
\hspace{-0.6em}
	\resizebox{0.75\textwidth}{!}{
		\setlength\tabcolsep{6pt}
		\renewcommand\arraystretch{1.1}
		\begin{tabular}{|c||rc||cc|}
			\hline\thickhline
			\rowcolor{mygray}
			\texttt{test} set &Method & Backbone & \texttt{top-1} & \texttt{top-5}  \\	\hline	\hline
			\multirow{2}{*}{MatchedFrequency} &ResNet \cite{he2016deep} &\multirow{2}{*}{ResNet50}  & 63.30\%& 84.70\%\\
             &\textbf{DNC}-ResNet& & \textbf{63.96}\% & \textbf{85.17}\%\\\hline
            \multirow{2}{*}{Threshold0.7} &ResNet \cite{he2016deep} &\multirow{2}{*}{ResNet50}  & 72.70\%& 92.00\%\\
             &\textbf{DNC}-ResNet& & \textbf{73.59}\% & \textbf{92.26}\%\\\hline
            \multirow{2}{*}{TopImages} &ResNet \cite{he2016deep} &\multirow{2}{*}{ResNet50}  & 78.10\%& 94.70\%\\
             &\textbf{DNC}-ResNet& & \textbf{78.62}\% & \textbf{94.96}\%\\\hline
	\end{tabular}}
	\label{table:ImageNetv2}
\end{table}

\section{Sinkhorn-Knopp \textit{vs} $k$-means clustering}
\label{sec_app:A5_add_3}
To further probe the impact of Sinkhorn-Knopp based clustering$_{\!}$~\cite{cuturi2013sinkhorn}, we further report the performance of DNC by using the classic k-means clustering algorithm, on CIFAR-10 and CIFAR100 datasets~\cite{CIFAR}. From Table~\ref{table:sinkhorn_kmeans} We can find that DNC with Sinkhorn-Knopp performs much better and is much more training-efficient.

\begin{table}[H]
\centering
\captionsetup{width=.99\textwidth}
	\caption{\textbf{Classification \texttt{top-1}$_{\!}$ and training time} on CIFAR-10 and CIFAR100 \cite{CIFAR} \texttt{test} sets.  See \S\ref{sec_app:A5_add_3} for more details.}
\hspace{-0.6em}
	\resizebox{0.99\textwidth}{!}{
		\setlength\tabcolsep{6pt}
		\renewcommand\arraystretch{1.1}
		\begin{tabular}{|c||rc||cc|}
			\hline\thickhline
			\rowcolor{mygray}
			 Dataset &Method & Backbone &\texttt{top-1} & Training time (hours)  \\	\hline	\hline
			\multirow{2}{*}{CIFAR10 \cite{CIFAR}} &\textbf{DNC}-$k$-means &\multirow{2}{*}{ResNet50}  & 93.88\%& 4.5\\
            &\textbf{DNC}-Sinkhorn&  & \textbf{95.78}\% & \textbf{2.5}\\\hline
            \multirow{2}{*}{CIFAR100 \cite{CIFAR}} &\textbf{DNC}-$k$-means &\multirow{2}{*}{ResNet50}  & 77.86\%& 16.3\\
            &\textbf{DNC}-Sinkhorn&  & \textbf{79.91}\% & \textbf{3.7}\\\hline
	\end{tabular}}
	\label{table:sinkhorn_kmeans}
\end{table}

\section{Comparison with Distance (Learning) based Classifiers}
\label{sec_app:A5_add_4}
We next compare our DNC with two metric based image classifiers~\cite{guerriero2018deepncm,wu2018improving} and one distance learning based segmentation model~\cite{wang2021exploring}.

We first compare DNC with~DeepNCM \cite{guerriero2018deepncm}. DeepNCM conducts similarity-based classification using class means.  As DeepNCM only describes its training procedures for CIFAR-10 and CIFAR-100 \cite{CIFAR}, we make the comparison on these two datasets to ensure fairness. From Table~\ref{table:metric_learning_1} we can find that, DNC significantly outperforms DeepNCM by +\textbf{2.11}\% on CIFAR-10 and +\textbf{7.16}\% on CIFAR-100, respectively.

DeepNCM simply abstracts each class into one single class mean, failing to capture complex
within-class data distribution. In contrast, DNC considers $K$ sub-centers per class. Note that this is not just increasing the number of class representatives. This requires accurately discovering the underlying data structure. DNC therefore jointly conducts automated online clustering (for mining sub-class patterns) and supervised representation learning (for cluster center based classification). Finding meaningful class representatives is extremely challenging and crucial for Nearest Centroids. As the experiment in \S\ref{sec_app:A5_add_3} and Table~\ref{table:sinkhorn_kmeans} revealed, simply adopting classic $k$-means causes huge performance drop and significant training speed delay.  Moreover, DeepNCM computes class mean on each batch, which makes poor approximation of the real class center. In contrast, DNC adopts the external memory for more accurate data densities modeling -- DNC makes clustering over the large memory of numerous training samples, instead of the relatively small batch. Thus DNC better captures complex within-class variants and addresses two key properties of prototypical exemplars: sparsity and expressivity \cite{quattoni2008transfer, biehl2009metric}, and eventually gains much more promising results.

\begin{table}[H]
\centering
\captionsetup{width=.7\textwidth}
	\caption{\textbf{Classification \texttt{top-1}$_{\!}$} on CIFAR-10 and CIFAR100 \cite{CIFAR} \texttt{test}. See \S\ref{sec_app:A5_add_4} for more details.}
\hspace{-0.6em}
	\resizebox{0.7\textwidth}{!}{
		\setlength\tabcolsep{8pt}
		\renewcommand\arraystretch{1.1}
		\begin{tabular}{|c||rc||c|}
			\hline\thickhline
			\rowcolor{mygray}
			Dataset &Method & Backbone & \texttt{top-1}   \\	\hline	\hline
			\multirow{2}{*}{CIFAR-10} &DeepNCM \cite{guerriero2018deepncm} &\multirow{2}{*}{ResNet50}  & 93.67\%\\
            &\textbf{DNC}& &  \textbf{95.78}\% \\\hline
            \multirow{2}{*}{CIFAR-100} &DeepNCM \cite{guerriero2018deepncm} &\multirow{2}{*}{ResNet50}  & 72.75\%\\
            &\textbf{DNC}&  & \textbf{79.91}\% \\\hline
	\end{tabular}}
	\label{table:metric_learning_1}
\end{table}

We also compare DNC with DeepNCA~\cite{wu2018improving}. DeepNCA is a deep $k$-NN classifier, which conducts similarity-based classification based on top-$k$ nearest training samples. As \cite{wu2018improving} only reports results on ImageNet~\cite{ImageNet}, we make the comparison on ImageNet to ensure fairness. Note that 130 training epochs are used, as in \cite{wu2018improving}. From Table~\ref{table:metric_learning_2} we can observe that, DNC outperforms DeepNCA.
As we discussed in \S\ref{sec:rw} and \S\ref{sec:cls}, DeepNCA poses huge storage demand, \ie, retaining the whole ImageNet training set (\ie, 1.2M images) to perform the $k$-NN decision rule, and suffers from very low efficiency, caused by extensive comparisons between each test sample and ALL the training images.  These limitations prevent the adoption of \cite{wu2018improving} in real application scenarios. In contrast, DNC only relies on a small set of class representatives (\ie, four sub-centroids per class) for decision-making and causes no extra computation budget during network deployment.

\begin{table}[H]
\centering
\captionsetup{width=.55\textwidth}
	\caption{\textbf{Classification \texttt{top-1}$_{\!}$} on ImageNet~\cite{ImageNet} \texttt{val}. See \S\ref{sec_app:A5_add_4} for more details.}
\hspace{-0.6em}
	\resizebox{0.55\textwidth}{!}{
		\setlength\tabcolsep{8pt}
		\renewcommand\arraystretch{1.1}
		\begin{tabular}{|rc||c|}
			\hline\thickhline
			\rowcolor{mygray}
			Method & Backbone &\texttt{top-1}   \\	\hline	\hline
			DeepNCA \cite{wu2018improving} &\multirow{2}{*}{ResNet50}   & 76.57\%\\
            \textbf{DNC}& & \textbf{76.64}\% \\\hline
	\end{tabular}}
	\label{table:metric_learning_2}
\end{table}

Finally, we compare DNC with ContrastiveSeg \cite{wang2021exploring} on Cityscapes \cite{cordts2016cityscapes} \texttt{val}, using ResNet101~\cite{he2016deep} backbone and DeepLabV3~\cite{chen2017rethinking} segmentation architecture. ContrastiveSeg applies contrastive learning to better shape the feature embedding space, so as to improve semantic segmentation performance. However, ContrastiveSeg still relies on the softmax classifier; it is essentially a distance learning boosted parametric classifier. From Table~\ref{table:metric_learning_2} we can observe that, DNC greatly surpasses ContrastiveSeg by 0.6\% \texttt{mIoU}.

\begin{table}[H]
\centering
\captionsetup{width=.7\textwidth}
		\caption{\textbf{Segmentation mIoU score} on Cityscapes \cite{cordts2016cityscapes} \texttt{val}. See \S\ref{sec_app:A5_add_4} for more details.}
\hspace{-0.6em}
	\resizebox{0.7\columnwidth}{!}{
		\setlength\tabcolsep{8pt}
		\renewcommand\arraystretch{1.05}
		\begin{tabular}{|rc||c|}
			\hline\thickhline
			\rowcolor{mygray}
			{Method~~~~~~} &Backbone & \texttt{mIoU} \\ \hline\hline
			ContrastiveSeg-DeepLab$_\text{V3}$ &\multirow{2}{*}{ResNet101}
			 & 79.1\% \\
			\textbf{DNC}-DeepLab$_\text{V3}$ & & \textbf{79.8\%} \\ \hline

	\end{tabular}}
	\label{table:metric_learning_3}
\end{table}

These three experiments solidly demonstrate our effectiveness on both image classification and segmentation tasks, even compared with other distance (learning) based counterparts.

\section{Additional Diagnostic Experiment}
\label{sec_app:A6}

\noindent\textbf{Output Dimensionality.} As stated in \S\ref{sec:exp}, the final output dimensionality of our DNC is set as many as the one of the last layer of the parametric counterpart, for the sake of fair comparison. However, owing to the distance-/similarity-based nature, DNC has the flexibility to handle any output dimensionality. In Table~\ref{table:feature_size}, we further study the influence of the output dimensionality of DNC. As seen, when setting the final output dimensionality as 1280, we can achieve \textbf{76.61\%} \texttt{top-1} acc., which is higher than the initial 2048 dimension configuration, \ie, 76.49\%. We attribute the reason for the better balance between memory capacity and feature dimensionality with the limitation of hardware computational budget -- when reducing the final output dimensionality, the expressibility of the final feature is weakened but more image features can be stored in the external memory for more accurate sub-centroid estimation.

\begin{table}[H]
	\centering
\captionsetup{width=.65\textwidth}
	\caption{{\textbf{Ablative experiments regarding the final output dimensionality} on ImageNet$_{\!}$~\cite{ImageNet}$_{\!}$ \texttt{val}. See \S\ref{sec_app:A6} for more details.}}
\hspace{-0.6em}
\renewcommand\arraystretch{1.1}
	\resizebox{0.50\textwidth}{!}{
			\setlength\tabcolsep{10pt}
			\begin{tabular}{|c||cc|}
				\hline\thickhline
				\rowcolor{mygray}
				\rowcolor{mygray}
                Output &\multicolumn{2}{c|}{ImageNet}\\
                \rowcolor{mygray}
				dimensionality & \texttt{top-1} & \texttt{top-5} \\
				\hline
				\hline
				640   & 76.23$\%$ &92.83$\%$ \\
				1024   & 76.28$\%$ &92.90$\%$ \\
                \textbf{1280}  & \textbf{76.61}$\%$ &\textbf{93.12}$\%$ \\
				2048   & 76.49$\%$ &93.08$\%$ \\
				\hline
			\end{tabular}
		}
	\label{table:feature_size}
\end{table}

\noindent\textbf{Temperature Parameter $\varepsilon$ in (\ref{eq:BIP2}).} Parameter $\varepsilon$ in (\ref{eq:BIP2}) trades off convergence speed with closeness to the original transport problem \cite{cuturi2013sinkhorn, asano2020self}. In Table~\ref{table:epsilon}, we further study the impact of $\varepsilon$ on ImageNet$_{\!}$~\cite{ImageNet}$_{\!}$ \texttt{val}.

\begin{table}[H]
	\centering
\captionsetup{width=.65\textwidth}
	\caption{\textbf{Ablative experiments regarding temperature parameter $\varepsilon$ in (\ref{eq:BIP2})} on ImageNet$_{\!}$~\cite{ImageNet}$_{\!}$ \texttt{val}. See \S\ref{sec_app:A6} for more details.}
\hspace{-0.6em}
\renewcommand\arraystretch{1.1}
	\resizebox{0.40\textwidth}{!}{
			\setlength\tabcolsep{10pt}
			\begin{tabular}{|c||cc|}
				\hline\thickhline
				\rowcolor{mygray}
                 &\multicolumn{2}{c|}{ImageNet}\\
                \rowcolor{mygray}
				\multirow{-2}{*}{$\epsilon$} & \texttt{top-1} & \texttt{top-5} \\
				\hline
				\hline
				0.01   & 76.34$\%$ &92.97$\%$ \\
                \textbf{0.05}  & \textbf{76.49}$\%$ &\textbf{93.08}$\%$ \\
				0.1   & 76.40$\%$ &93.02$\%$ \\
				\hline
			\end{tabular}
		}
	\label{table:epsilon}
\end{table}


\noindent\textbf{Number of Centroids $K$.} As shown in Table~\ref{table:kvalue_varies}, we set $K$ with different values based on the number of training samples of each class. Specifically, ImageNet contains between 732 and 1300 training images (\#images) per class. Then, $K=1$ is assigned to the class having between 732 and 874 training samples, $K=2$ to the class having between 875 and 1016 samples, $K=3$ to the class having between 1017 and 1158 samples, and $K=4$ to the class having between 1159 and 1300 samples. We can find that we gain slightly better performance, +0.06\% higher in \texttt{top-1} acc. when compared with fixing $K=4$ for all the classes.

\begin{table}[H]
	\centering
\captionsetup{width=.66\textwidth}
	\caption{\textbf{Ablative experiments with varying $K$ on different classes} on ImageNet$_{\!}$~\cite{ImageNet}$_{\!}$ \texttt{val}. See \S\ref{sec_app:A6} for more details.}
\hspace{-0.6em}
\renewcommand\arraystretch{1.1}
	\resizebox{0.55\textwidth}{!}{
			\setlength\tabcolsep{10pt}
			\begin{tabular}{|c||cc|}
				\hline\thickhline
				\rowcolor{mygray}
                $K$ &\multicolumn{2}{c|}{ImageNet}\\
                \rowcolor{mygray}
				range & \texttt{top-1} & \texttt{top-5} \\
				\hline
				\hline
				unique value 4   & 76.49$\%$ &93.08$\%$ \\
				varying between 1 and 4   & \textbf{76.55}$\%$ &\textbf{93.10}$\%$ \\
				\hline
			\end{tabular}
		}
\vspace{-16pt}
	\label{table:kvalue_varies}
\end{table}

\section{Memory Cost}
\label{sec_app:A7}
\vspace{-8pt}
In our experiment, we only adopt external memory for ImageNet classification. Below we provide more discussion regarding this point. Semantic segmentation is a pixel-wise classification task, where each training image provides numerous pixel samples for each class. For ImageNet classification, however, each training image is only assigned to one single class. Moreover, ImageNet has 1K classes, while in general semantic segmentation only has dozens of classes. Therefore, for a training mini-batch of, for example, 256 images, every class in segmentation usually has many training pixel samples in each mini-batch; this allows us to use a large $K$ for clustering. However, under the same setting, for each mini-batch, there must have many ImageNet classes that do not have corresponding image samples -- we have 1000 classes but each mini-batch only has 256 training images. This is why we need to build an external memory during ImageNet classification. This is also why applying Nearest Centroids for batch-wise ImageNet classification training is extremely challenging; \cite{guerriero2018deepncm} cannot handle ImageNet classification, as it only computes class means in a batch-wise manner.

Table~\ref{table:memory_withK} and \ref{table:memory_withBatch} provide statistics of GPU memory cost (per GPU usage) with respect to the number of class centroids $K$ and external memory size respectively. The statistics are gathered during the training of {DNC}-ResNet50 on ImageNet, using eight V100 GPUs. In our experiment, we set the size of the external memory  as 256,000 image samples (\ie, 1000 batches) and $K\!=\!4$ for DNC-ResNet50 (\ie, a total of 4000 class sub-centroids on ImageNet). More specifically, for a memory with \#batch=1000, it stores 8 \#gpu $\times$ 32 \#batch size $\times$ 1000 \#batch = 256000 \#image examples. In comparison, for segmentation, when we set $K\!=\!10$ (\ie, a total of 1500 class sub-centroids for ADE20K), we have 65,536 pixel training samples in each mini-batch of 16 training images, without using memory.

\begin{table}[H]
	\caption{\textbf{Statistics of GPU memory cost} with respect to the number of class centroids $K$ and external memory size, where 1000 batches = 256000 image samples. See \S\ref{sec_app:A7} for more details.}
\vspace{-7px}
\centering
		\hspace{-0.7em}
    \begin{subtable}{0.475\linewidth}
		\resizebox{\textwidth}{!}{
			\setlength\tabcolsep{3.5pt}
			\begin{tabular}{|c||c|}
				\hline\thickhline
				\rowcolor{mygray}
                $K$ (Fixed memory size:  &GPU memory \\
                \rowcolor{mygray}
				1000 batches) &cost (GB per GPU) \\
				\hline
				\hline
				1   & 16.07  \\
				2   & 19.04  \\
				3   & 22.03  \\
				4   & 26.31  \\
				\hline
			\end{tabular}
		}
		\vspace{-0px}
		\setlength{\abovecaptionskip}{0.4cm}
		\setlength{\belowcaptionskip}{-0.1cm}
		\caption{GPU memory cost \textit{w.r.t} number of centroids  $K$}
		\vspace{-6px}
		\label{table:memory_withK}
		\end{subtable}
	\begin{subtable}{0.485\linewidth}
		\resizebox{\textwidth}{!}{
			\setlength\tabcolsep{14pt}
			\begin{tabular}{|c||c|}
				\hline\thickhline
				\rowcolor{mygray}
				\rowcolor{mygray}
                memory size &GPU memory \\
                \rowcolor{mygray}
				$_{\!\!}$(Fixed $K=4$)$_{\!\!}$ &cost (GB per GPU) \\
				\hline
				\hline
				400 batches   & 11.88  \\
				600 batches  & 16.35  \\
				800 batches & 21.65  \\
				1000 batches & 26.35  \\
				\hline
			\end{tabular}
		}
		\vspace{-0px}
		\setlength{\abovecaptionskip}{0.4cm}
		\setlength{\belowcaptionskip}{-0.1cm}
		\caption{GPU memory cost \textit{w.r.t} external memory size}
		\vspace{-6px}
		\label{table:memory_withBatch}
		\end{subtable}
\vspace{-14pt}
\end{table}

\FloatBarrier
\section{Additional  Study  of Ad-hoc Explainability}
\label{sec_app:A8}
\begin{figure*}[t]
\vspace{-2pt}
  \centering
      \includegraphics[width=1 \linewidth]{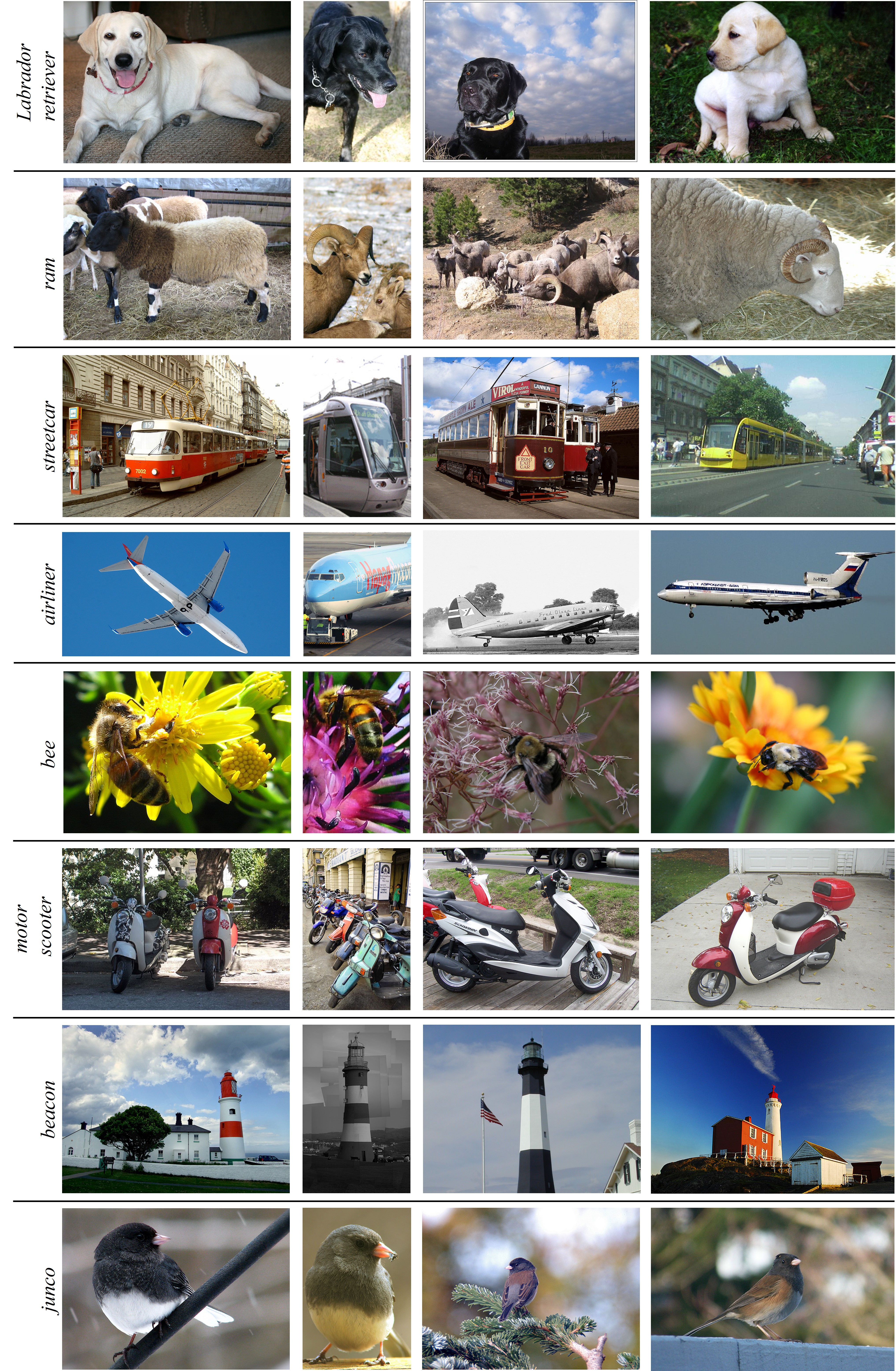}
     \vspace{-17pt}
\caption{Sub-centroid images for eight randomly chosen classes  from ImageNet \cite{ImageNet}. See \S\ref{sec_app:A8} for more details.}
\label{fig:vsx}
\vspace{-13pt}
\end{figure*}

\vspace{-8pt}
\noindent\textbf{Interpretable Class Sub-centroids.} In Fig.~\ref{fig:vsx}, we show more examples of sub-centroid images for eight ImageNet classes. These representative images are automatically discovered by DNC, and can be intuitively viewed by users. As seen, the class sub-centroid  images are able to capture diverse characteristics of their classes, in the aspects of appearance, viewpoints, scales, illuminations, \textit{etc}.

\begin{figure*}[t]
  \centering
       \vspace{-10pt}
      \includegraphics[width=0.7\linewidth]{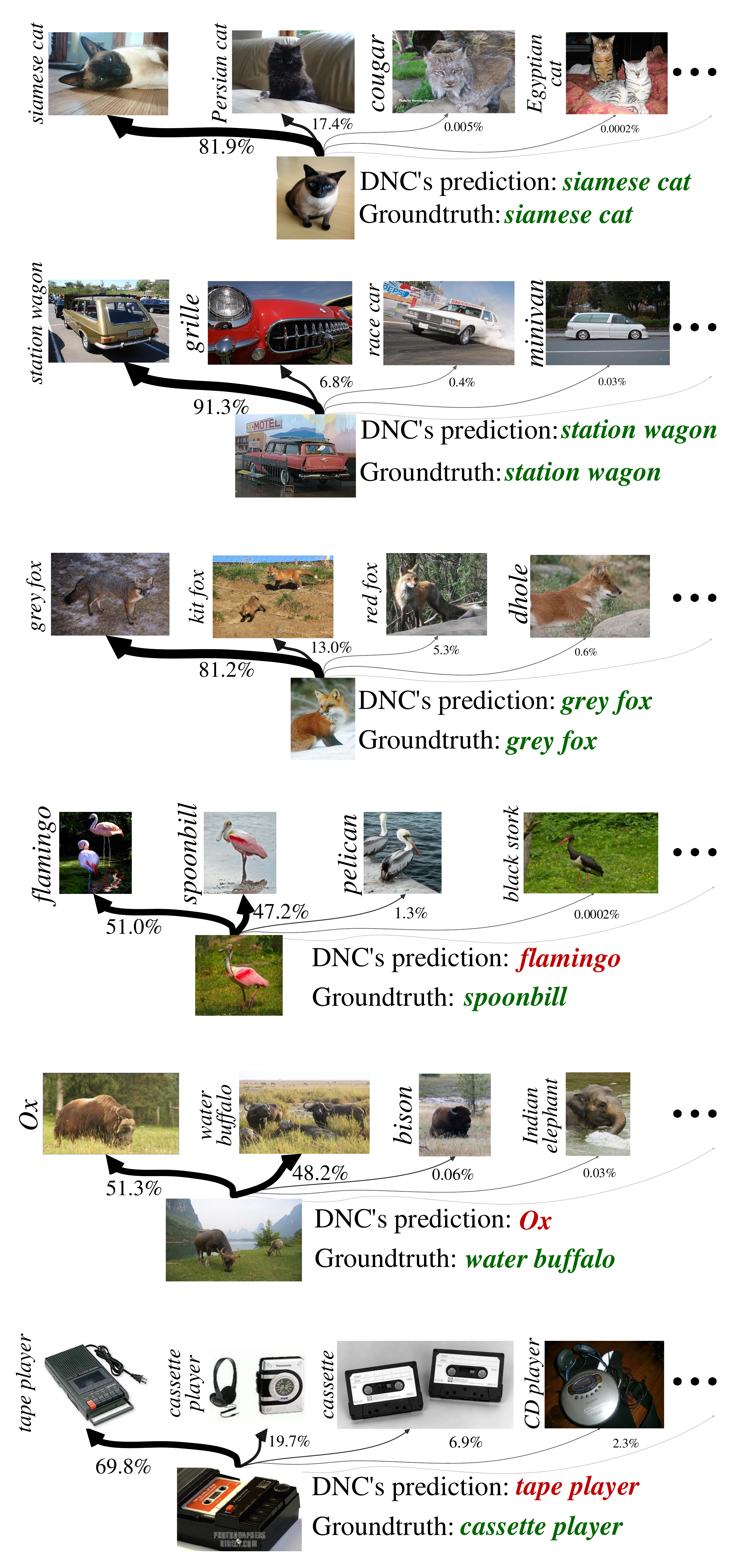}
     \vspace{-7pt}
    \setlength{\abovecaptionskip}{15pt}
\caption{More examples on DNC interpreting its predictions based on its computed similarity to class sub-centroid images. For each test image, we plot the normalized similarities for the corresponding closest sub-centroids from the top-4 scoring classes. See \S\ref{sec_app:A8} for more details.}
\label{fig:vex}
\vspace{-13pt}
\end{figure*}

\noindent\textbf{Interpret Prediction Based on (Dis)similarity to Sub-centroid Images.} Fig.~\ref{fig:vex} provides more results regarding the (dis)similarity-based interpretation of DNC prediction. As seen, based on the similarity of test images to the class sub-centroid images, users can clearly understand the decision making mode and make verification. DNC's compelling explainability enables it to establish trustworthiness with humans and empowers its potential in high stake applications.

\FloatBarrier
\section{Qualitative Results on Semantic Segmentation}
\label{sec_app:A9}

\begin{figure*}[t]
  \centering
      \includegraphics[width=1 \linewidth]{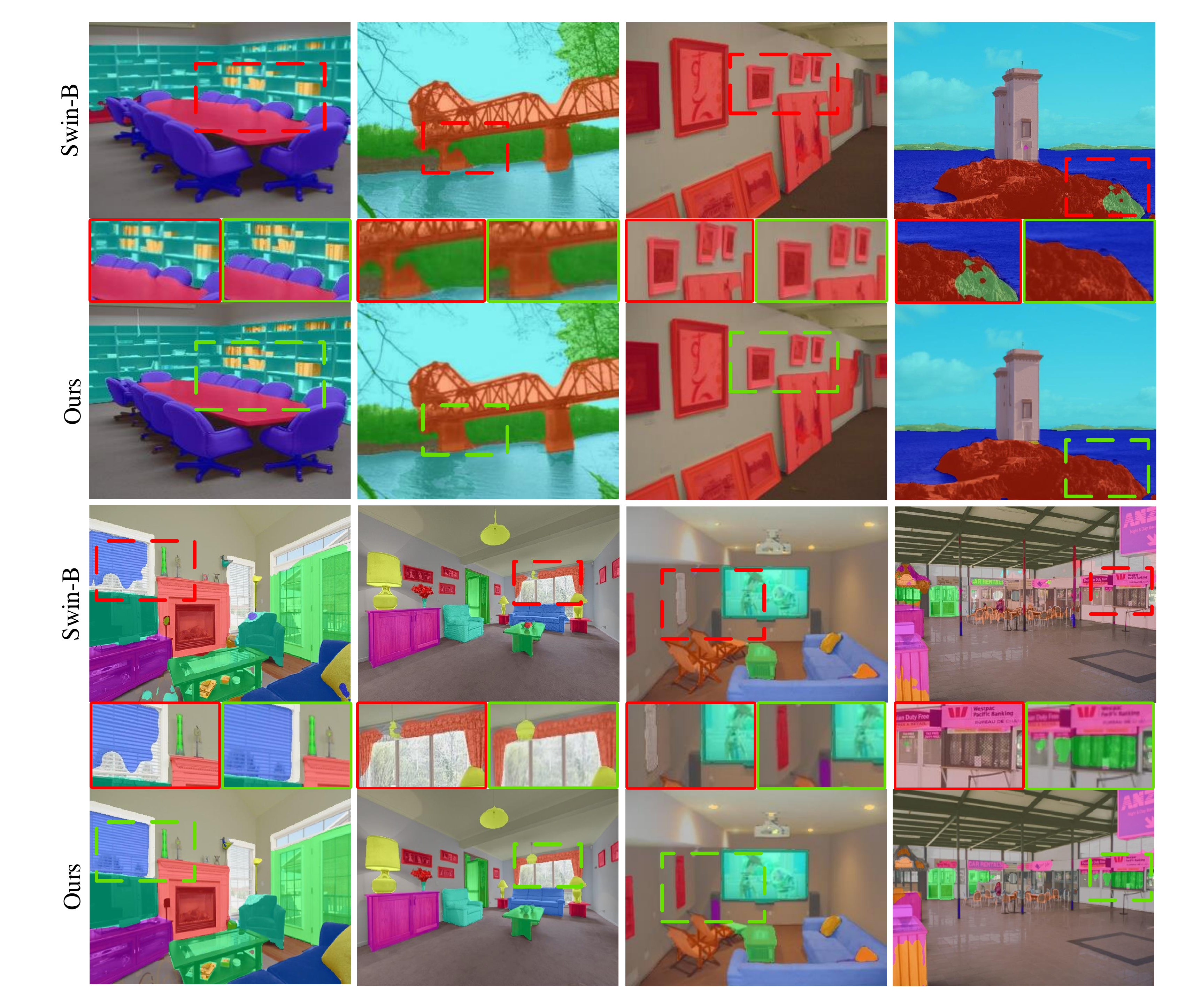}
     \vspace{-17pt}
     \setlength{\abovecaptionskip}{5pt}
\caption{\textbf{Qualitative semantic segmentation results} of UperNet$_{\!}$~\cite{xiao2018unified}-Swin-B \cite{liu2021swin} and DNC on ADE20K \cite{zhou2017scene} \texttt{val}. Red and green bounding boxes represent the same zoom-in area on UperNet$_{\!}$~\cite{xiao2018unified}-Swin-B \cite{liu2021swin} and DNC, respectively. See \S\ref{sec_app:A9} for more details.}
\label{fig:ade20k}
\vspace{-13pt}
\end{figure*}

\begin{figure*}[t]
  \centering
      \includegraphics[width=\linewidth]{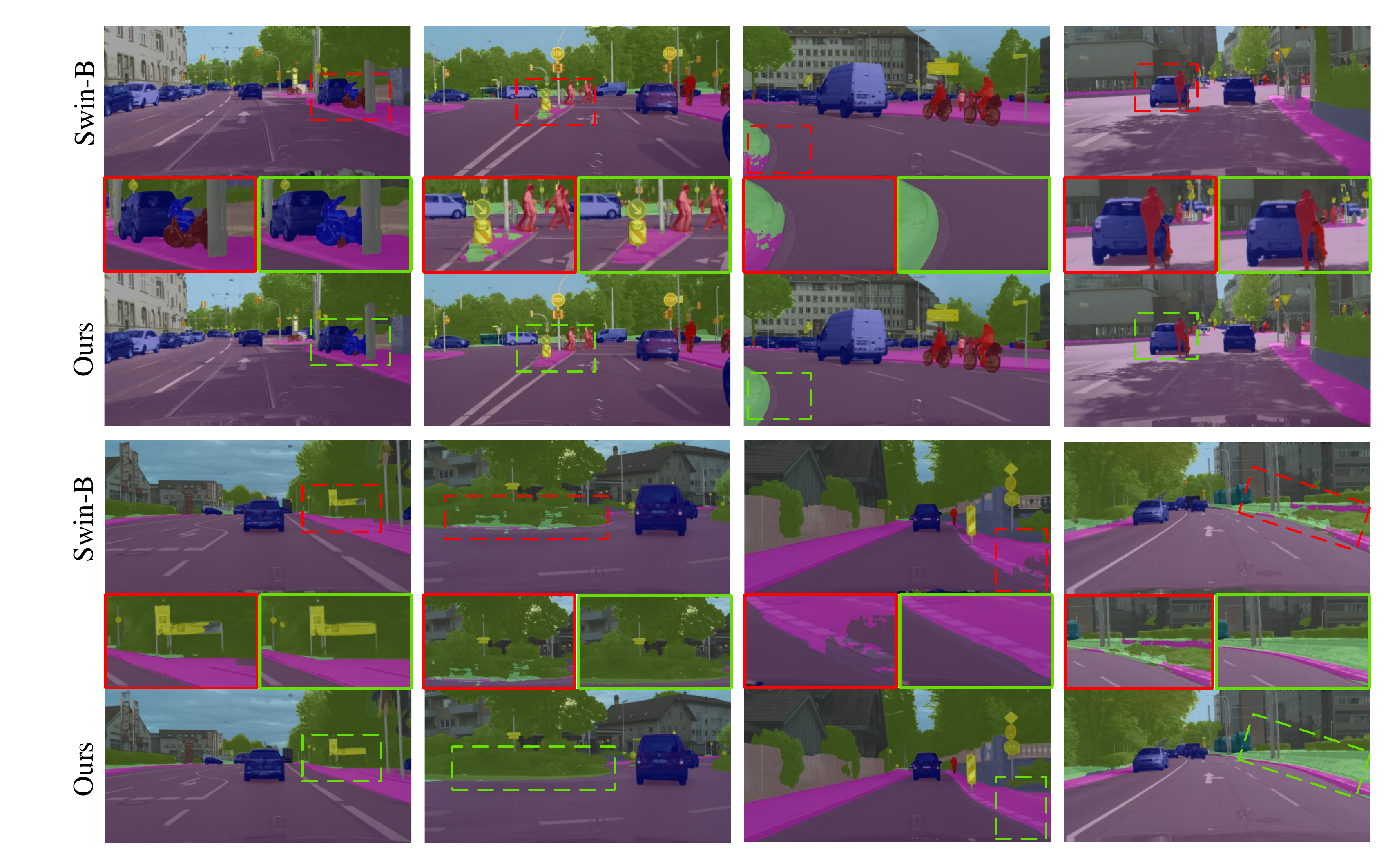}
     \vspace{-17pt}
     \setlength{\abovecaptionskip}{5pt}
\caption{\textbf{$_{\!}$ Qualitative$_{\!}$  semantic$_{\!}$  segmentation$_{\!}$  results}$_{\!}$  of$_{\!}$  UperNet$_{\!}$~\cite{xiao2018unified}-Swin-B$_{\!}$~\cite{liu2021swin}$_{\!}$  and$_{\!}$  DNC$_{\!}$  on$_{\!}$  Cityscapes$_{\!}$~\cite{cordts2016cityscapes}$_{\!}$  \texttt{val}. Red and green bounding boxes represent the same zoom-in area on UperNet$_{\!}$~\cite{xiao2018unified}-Swin-B \cite{liu2021swin} and DNC, respectively. See \S\ref{sec_app:A9} for more details.}
\label{fig:cityscapes}
\vspace{-13pt}
\end{figure*}

Fig.~\ref{fig:ade20k} and Fig.~\ref{fig:cityscapes} illustrate a few representative visual examples of semantic segmentation results on ADE20K~\cite{zhou2017scene} and Cityscapes~\cite{cordts2016cityscapes}, respectively. {In comparison with the parametric counterpart, our approach produces more precise segments in different challenging scenes (for example, where objects with drastic photometric or geometric appearances). For instance, UperNet$_{\!}$~\cite{xiao2018unified}-Swin-B \cite{liu2021swin} confuses on neighbouring objects with similar colors ($e.g.$, desk and chair) and leaves a large false-negative regions (see the first image of Fig.~\ref{fig:ade20k}); it also has difficulties in segmenting small scale objects ($e.g.$, motorbike semantic parsing).
Among these examples, DNC consistently demonstrates supreme performance.} {Essentially, we argue that the proposed DNC has a stronger ability to supervises the pixel embedding space via anchoring sub-centroids directly, leading to a better predictions on segmenting such hard cases.}

\section{Additional Literature Review}
\label{sec_app:B}
This section gives additional review of representative literature on metric-/distance-learning and clustering-based unsupervised representation learning.

\noindent\textbf{Metric Learning.} The goal of metric learning (also a.k.a distance learning) is to learn a distance metric/embedding which brings together similar samples and pushes away dissimilar ones. Metric learning has a long history, dating back to some early work for more than few decades ago \cite{short1981optimal,hastie1995discriminant}. In particular, diverse metric learning objective functions, such as contrastive loss \cite{bromley1993signature, hadsell2006dimensionality, khosla2020supervised}, triplet loss \cite{schroff2015facenet}, quadruplet loss \cite{chen2017beyond}, and $n$-pair loss$_{\!}$ \cite{sohn2016improved}, were proposed to measure similarity in the feature space for representation learning, and showed significant benefit in a wide range of applications, such as image retrieval \cite{wu2017sampling}, face recognition \cite{schroff2015facenet, taigman2014deepface, wang2018cosface, cao2022towards, cao2021vector}, and person re-identification \cite{xiao2017joint}, to name a few representative ones. Recently, metric learning gained astonishing success in learning transferable deep representations from massive unlabeled data~\cite{kaya2019deep}. A family of \textit{instance-based} approaches
used the contrastive loss \cite{gutmann2010noise,oord2018representation} to explicitly compare pairs of image representations~\cite{oord2018representation,hjelm2019learning,chen2020improved,chen2020simple,he2020momentum}. Another group of methods adopted a \textit{clustering-based} strategy; they learn unsupervised representations by discriminating
between groups of images without expensive pairwise comparison between image instances \cite{cliquecnn2016,xie2016unsupervised,caron2018deep,yan2020clusterfit,caron2020unsupervised,li2020prototypical,asano2020self,van2020scan,tao2021idfd}. More recently, there are some efforts that revisit the idea of metric learning in supervised learning setting~\cite{khosla2020supervised,sun2020circle,suzuki2019arc,wang2021exploring}.

As distance-/similarity-based classifiers rely on the similarity between samples and class representatives for classification,  the fields of metric learning and distance-based classification are naturally related and the selection of a proper distance measure impacts the success of distance-based classifiers$_{\!}$~\cite{biehl2013distance}. Historically, metric learning and class center discovery are two critical research topics in the field of distance-based classification. As a nonparametric, distance-based classifier, DNC can be viewed as a learnable metric function, which is trained to compare data samples under the guidance of the corresponding semantic labels. Although current distance learning based algorithms also optimize the feature space by comparing data samples, they need parametric softmax for classification. Their trained models are still black-box parametric classifiers without any interpretability. In sharp contrast, DNC directly assigns an observation to the class of the closet centroids, without using parametric softmax. Moreover, its distance-based classification decision-making mode allows DNC to effortless adopt existing metric learning techniques (and the way of its current training can be already viewed as performing metric learning).

\noindent\textbf{Clustering-based Self-supervised Representation Learning.} There is a recent trend to bind self-supervised representation learning with clustering. Basically, clustering-based self-supervised representation learning is more efficient for large-scale training data and more tolerant of the similarity (semantic structure) among data samples, compared with the instance-level counterpart. More specifically, early approaches \cite{cliquecnn2016,caron2018deep,ji2019invariant,zhan2020online,yan2020clusterfit,asano2020self,van2020scan} learn representations of image samples and cluster assignments in an \textit{alternative} manner, \ie, group features into clusters to derive pseudo supervisory signal and subsequently employ it for supervising representation learning.  In very recent, numerous efforts have been devoted to \textit{simultaneous} clustering and representation learning based on, \eg, data reconstruction \cite{xie2016unsupervised,yang2017towards}, mutual information maximization \cite{ji2019invariant,wu2019deep,tschannen2019mutual}, or contrastive instance discrimination \cite{wu2018unsupervised,tao2021clustering,tsai2020mice,caron2020unsupervised,li2020prototypical,tao2021idfd}.

Our work is also related to these clustering based unsupervised representation learning methods, especially the ones~\cite{asano2020self,caron2020unsupervised} resorting to the fast Sinkhorn-Knopp algorithm~\cite{cuturi2013sinkhorn} for robust clustering. They aim to \textit{learn transferable representation} from massive \textit{unlabeled} data. Although also involving a similar clustering procedure for automatic sub-pattern mining, DNC targets at building a strong \textit{similarity-based classification network} in the standard \textit{supervised learning} setting. In DNC, the automatically discovered class sub-centroids are informative class representatives, which explicitly capture latent data structure of each class, and serve as classification evidence with clear physical meaning. The whole training procedure is a hybrid of class-wise online clustering (for unsupervised sub-class discovery) and sub-centroid based classification (for supervised representation learning). This well addresses the nature of Nearest Centroids and brings novel insights into the visual recognition task itself.

\section{Limitation and Future Work}
\label{sec_app:C}
\noindent\textbf{Limitation.} {One limitation of our approach is that the Sinkhorn-Knopp algorithm runs in time $\widetilde{O}(\frac{n^{2}}{\epsilon^{3}})$ which would reduce the training efficiency. Though in practice, we find 3 sinkhorn loops per training iteration is sufficient enough for model representation, bringing a minor computational overhead ($i.e.$, $\sim$5\% training delay on ImageNet). This also indicates possible directions for our future research.}

{
\noindent\textbf{Social Impact.} This work introduces DNC possessing the nature of nested simplicity, intuitive decision-making mechanism and even \textit{ad-hoc} explainability. On positive side, the approach advances model accuracy and is valuable in safety-sensitive applications by showing the advanced robustness on sub-categories discovery, $e.g.$, quality analytics, autonomous driving~\cite{liu2020video, wang2019end, liu2021visual}, \textit{etc.}. For potential negative social impact, our DNC struggles in handling out-of-distribution data, which is a common limitation of all the discriminative classifiers. Hence its utility in open-world scenarios should be further examined.
}

\noindent\textbf{Future Work.} Despite DNC's systemic simplicity and efficacy, it also comes with new challenges and unveils some intriguing questions. For example, incorporating more powerful, time-efficient online clustering algorithms into DNC might improve training speed and test accuracy. Also, the number of class centroids $K$ currently is set to a fixed value for all classes, which may not be optimal given that intra-class variability varies across classes. Our experiments in \S\ref{sec_app:A6} and Table~\ref{table:kvalue_varies} also suggest that simply varying $K$ with the number of training samples of the class can boost performance. Thus adopting the clustering algorithms that do not require a predefined and fixed number of clusters~\cite{ronen2022deepdpm} may allow DNC to automatically determine $K$ for different classes, which eventually benefit performance. In addition, instead of only considering first-order statistics, DNC could be enhanced by second-order statistics, which contain more useful information, but must contend with the computational overhead they impose. Another essential future direction deserving of further investigation is the in-depth analysis of the intrinsic properties of DNC, such as its robustness against perturbation, adversarial attack~\cite{cheng2022physical, cheng2023adversarial}, and out-of-distribution data, with the comparison of the softmax based counterpart. This endeavor would help us to better understand the nature of parametric and nonparametric classifiers and reveals directions for further improvement. Furthermore, we will explore the possibility of unifying close-set and open-world visual recognition within$_{\!}$ our$_{\!}$ framework. Finally, considering the similarity-/distance-based nature of DNC, the  incorporation of metric learning based training objectives is also another promising direction for further boosting the performance. Given the vast number of technique breakthroughs in recent years, we expect a flurry of innovation towards these promising directions. Overall, we believe the results presented in this paper warrant further exploration.

\end{document}